\documentclass{article} % For LaTeX2e

\usepackage[accepted]{icml2025}

\usepackage{microtype}
\usepackage{hyperref}

\usepackage{mathtools}
\usepackage{paralist}
\usepackage{graphicx} 
\usepackage{booktabs} % For \Xhline and better table lines
\usepackage{color, xcolor}
\usepackage{multirow}
\usepackage{booktabs}
\usepackage{graphicx}
\usepackage{pifont} % 用于打勾和打叉
\usepackage{xcolor} % 用于颜色
\usepackage{array} % 可选，用于增强表格功能
\usepackage{caption}
\usepackage{tabularx} % 用于自动调整列宽
\usepackage{etoc}        % 用于自定义目录
\usepackage{appendix}    % 可选，用于增强附录处理
\usepackage{tocbibind}   % 可选，用于自动将附录添加到目录
\usepackage{titlesec}    % 可选，用于自定义标题格式
\usepackage{hyperref}
\usepackage{url}
\usepackage{cancel}
\usepackage{caption}
\usepackage{subcaption}
\usepackage{tcolorbox}
\usepackage{makecell}
\usepackage{hyperref}
\usepackage{amsfonts,amsmath,amssymb,amsthm} 
\usepackage{algorithm}
\usepackage[commentColor=black,beginLComment=/*~, endLComment=~*/]{algpseudocodex}

\usepackage[capitalize,noabbrev]{cleveref}
\usepackage{enumitem}
\usepackage{amsmath,amsfonts,bm}

\def\eqref#1{equation~\ref{#1}}
\def\1{\bm{1}}

\def\va{{\bm{a}}}

\def\ve{{\bm{e}}}

\def\vu{{\bm{u}}}

\def\vx{{\bm{x}}}

\def\vz{{\bm{z}}}

\def\mA{{\bm{A}}}

\def\mE{{\bm{E}}}

\def\mI{{\bm{I}}}

\def\mP{{\bm{P}}}

\def\mU{{\bm{U}}}

\def\mW{{\bm{W}}}
\def\mX{{\bm{X}}}

\def\mZ{{\bm{Z}}}

\def\mPhi{{\bm{\Phi}}}

\DeclareMathAlphabet{\mathsfit}{\encodingdefault}{\sfdefault}{m}{sl}
\SetMathAlphabet{\mathsfit}{bold}{\encodingdefault}{\sfdefault}{bx}{n}

\def\calN{{\mathcal{N}}}

\newcommand{\E}{\mathbb{E}}

\newcommand{\R}{\mathbb{R}}

\DeclareMathOperator*{\argmin}{arg\,min}

\newcommand{\zz}[1]{\textcolor{blue}{ [{\em ZZ:} #1]}}

\definecolor{mygreen}{RGB}{0, 150, 0}
\newcommand{\mh}[1]{\textcolor{mygreen}{[{\em mh:} #1]}}
\newcommand{\cmark}{\textcolor{mygreen}{\ding{51}}}%
\newcommand{\xmark}{\textcolor{red}{\ding{55}}}%

\theoremstyle{plain}
\newtheorem{theorem}{Theorem}[section]
\newtheorem{proposition}[theorem]{Proposition}
\newtheorem{lemma}[theorem]{Lemma}

\theoremstyle{definition}
\newtheorem{definition}[theorem]{Definition}

\theoremstyle{remark}
\newtheorem{remark}[theorem]{Remark}

\etocdepthtag.toc{mtchapter}
\etocsettagdepth{mtchapter}{subsection}
\etocsettagdepth{mtappendix}{none}

\icmltitlerunning{Analyzing and Mitigating Model Collapse in Rectified Flow Models}
\begin{document}

\twocolumn[
\icmltitle{Analyzing and Mitigating Model Collapse in Rectified Flow Models}
\icmlsetsymbol{equal}{*}

\icmlsetsymbol{equal}{*}

\begin{icmlauthorlist}
\icmlauthor{Huminhao Zhu}{osu}
\icmlauthor{Fangyikang Wang}{ind}
\icmlauthor{Tianyu Ding}{comp}
\icmlauthor{Qing Qu}{um}
\icmlauthor{Zhihui Zhu}{osu}
%\icmlauthor{}{sch}
% \icmlauthor{Firstname8 Lastname8}{sch}
% \icmlauthor{Firstname8 Lastname8}{yyy,comp}
%\icmlauthor{}{sch}
%\icmlauthor{}{sch}
\end{icmlauthorlist}

\icmlaffiliation{osu}{The Ohio State University}
\icmlaffiliation{ind}{Independent Researcher}
\icmlaffiliation{comp}{Microsoft}
\icmlaffiliation{um}{University of Michigan, Ann Arbor}

\icmlcorrespondingauthor{Zhihui Zhu}{zhu.3440@osu.edu}
% \icmlcorrespondingauthor{Firstname2 Lastname2}{first2.last2@www.uk}

% You may provide any keywords that you
% find helpful for describing your paper; these are used to populate
% the "keywords" metadata in the PDF but will not be shown in the document
\icmlkeywords{Generative Models, ICML}

% \vskip 0.3in
% ]

{\vspace{0.75em}%
\centering
\includegraphics[width=0.9\linewidth]{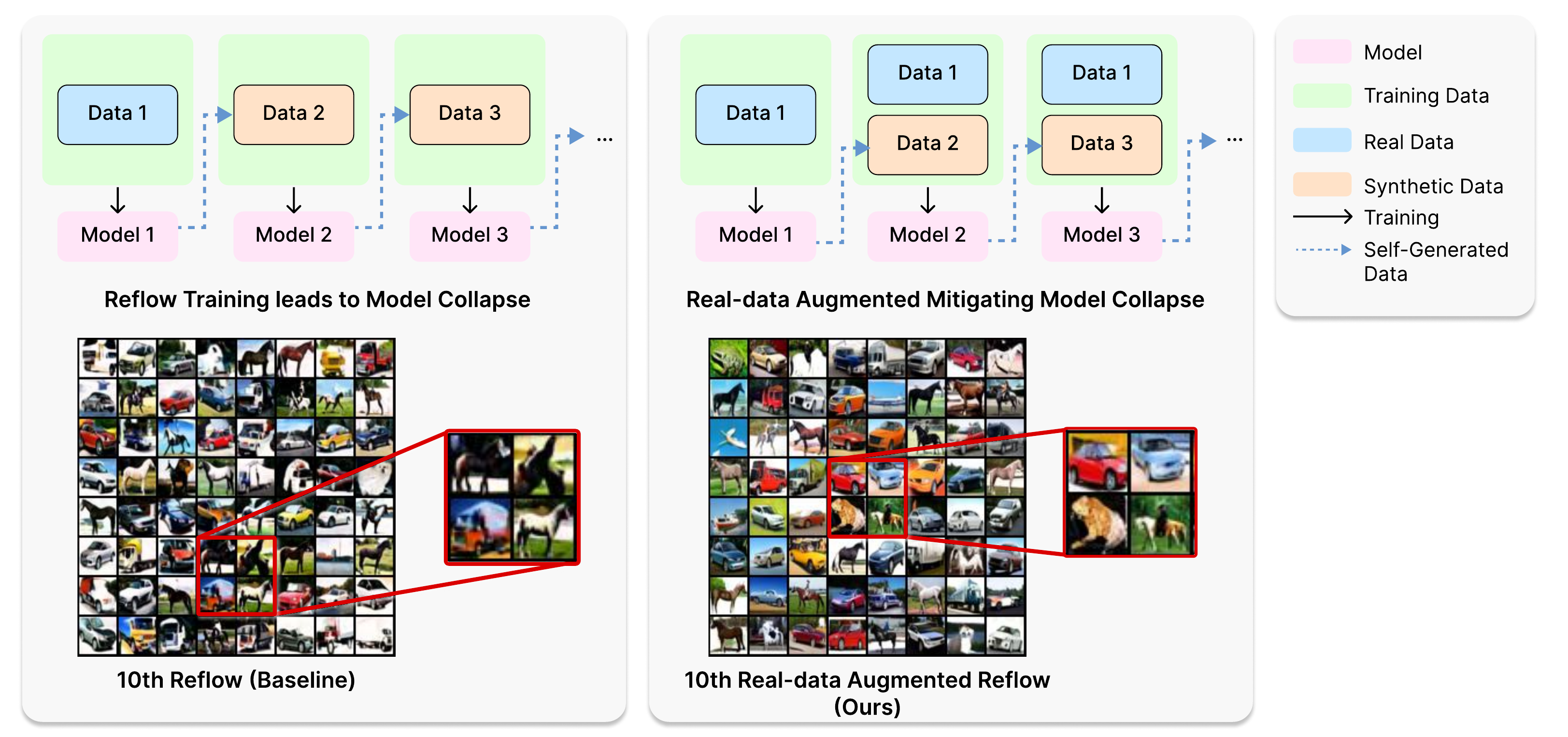}
% \vspace{-1em}
\captionof{figure}{\textbf{Two Scenarios for Studying Model Collapse.} 
\textcolor{blue}{Top}: Rectified Flow uses the Reflow method, iteratively relying on self-generated data to straighten the flow and improve sampling efficiency.
\textcolor{red}{Left}: Model collapse occurs when models are repeatedly trained on their own outputs, progressively degrading performance.
\textcolor{red}{Right}: Adding real data at each iteration mitigates collapse by preserving sample quality.
\textcolor{blue}{Bottom}: Visualization of correction streams after 10 iterations. The baseline lacks color and produces blurry, mixed outputs, whereas our approach maintains clarity and fidelity.
% \zz{change Reflow-RCA to RA Reflow}
% \textcolor{blue}{Top:} Model Collapse occurs when successive iterations of generative models, trained on their own outputs, progressively degrade in performance, ultimately rendering the final model ineffective. \textcolor{red}{Left:} Illustrates MC by replacing data with each training iteration. \textcolor{red}{Right:} Depicts the scenario where original real data is added at each iteration, demonstrating that incorporating real data prevents the model from collapsing. \textcolor{blue}{Bottom:} The correction streams trained in both modes after 10 iterations. The baseline lacks color, and the images are blurry and mixed. With our method, the images maintain their generated quality. 
% \zz{All the content in the caption is already known; here the unique thing is about reflow. So we need to highlight reflow. Any particular reason to put it before abstract?} \zh{If we want to highlight Reflow, then Figure 2 is the best, but Figure 2 requires a lot of prior knowledge. I think the image before abstract can only provide a better understanding. I will try to emphasize Reflow in the image.}\zz{I mean in the caption description, not the image.}
\vspace{-0.75em}
}  
    % \vspace{-5mm}
  \label{fig:method}
% \label{fig:teaser}
}

\vskip 0.3in
]
% \vskip 0.3in
% % \maketitle
% \begin{figure*}[ht]
%   \centering
% \end{figure*}

\printAffiliationsAndNotice{}  % leave blank if no need to mention equal contribution

\begin{abstract}
Training with synthetic data is becoming increasingly inevitable as synthetic content proliferates across the web, driven by the remarkable performance of recent deep generative models. This reliance on synthetic data can also be intentional, as seen in Rectified Flow models, whose Reflow method iteratively uses self-generated data to straighten the flow and improve sampling efficiency. However, recent studies have shown that repeatedly training on self-generated samples can lead to {\it model collapse (MC)}, where performance degrades over time. Despite this, most recent work on MC either focuses on empirical observations or analyzes regression problems and maximum likelihood objectives, leaving a rigorous theoretical analysis of reflow methods unexplored. In this paper, we aim to fill this gap by providing both theoretical analysis and practical solutions for addressing MC in diffusion/flow models. We begin by studying Denoising Autoencoders (DAEs) and prove performance degradation when DAEs are iteratively trained on their own outputs. To the best of our knowledge, we are the first to rigorously analyze model collapse in DAEs and, by extension, in diffusion models and Rectified Flow. Our analysis and experiments demonstrate that rectified flow also suffers from MC, leading to potential performance degradation in each reflow step. Additionally, we prove that incorporating real data can prevent MC during recursive DAE training, supporting the recent trend of using real data as an effective approach for mitigating MC. Building on these insights, we propose a novel {\it Real-data Augmented Reflow (RA Reflow)} and a series of improved variants, which seamlessly integrate real data into Reflow training by leveraging reverse flow. Empirical evaluations on standard image benchmarks confirm that RA Reflow effectively mitigates model collapse, preserving high-quality sample generation even with fewer sampling steps.

\end{abstract}

\vspace{-5mm}
\section{Introduction}
% \zz{need to mention \Cref{fig:method} somewhere.}
Generative modeling aims to produce synthetic data that is indistinguishable from genuine data distributions. While deep generative models have achieved remarkable success across images, audio, and text \citep{rombach2022high, ramesh2022hierarchical, chen2020wavegrad, achiam2023gpt, touvron2023llama}, the increasing volume of synthetic data introduces significant challenges. As shown in \Cref{fig:method}, a critical issue is \textit{model collapse (MC)}, where generative models trained iteratively on their own outputs progressively degrade in performance \citep{shumailov2023curse}. This degradation not only affects the quality of generated data but also poses risks when synthetic data is inadvertently included in training datasets, leading to self-consuming training loops \citep{alemohammad2023selfconsuming,briesch2023large}. Most recent work on model collapse centers on empirical observations \citep{shumailov2023curse, alemohammad2023selfconsuming}, and most of the theoretical study addresses regression problems \citep{fu2024towards, dohmatob2024model, gerstgrasser2024model} or maximum likelihood objectives \citep{bertrand2023stability}, leaving a rigorous theoretical analysis of diffusion/flow models unexplored.

% \zz{we need to briefly summarize the recent development on MC somewhere: empirical observation, lacks theoretical analysis (most focuses on regression problems)}

%Simulation-free models and their variants—such as diffusion models \citep{song2019generative, song2020score, ho2020denoising}, flow matching \citep{lipman2022flow, pooladian2023multisample, tong2023improving}, and Rectified Flow \zz{Rectified Flow or rectified flow?} \citep{liu2022flow}—have drawn increasing attention. Among these models, Rectified Flow stands out due to its rapid development and extensive foundational and large-scale work \citep{esser2024scaling}. Unlike typical diffusion models, rectified flow exploits a reflow procedure that iteratively utilizes self-generated data as training data to straighten the flow and improve sampling efficiency, which closely aligns with the definition of MC. This direct use of self-generated data makes Rectified Flow an ideal candidate for studying and addressing MC. However, previous studies on Rectified Flow have primarily focused on scaling up the model or applying distillation techniques \citep{lee2024improving, liu2023instaflow, esser2024scaling}, while neglecting a thorough analysis of MC itself. Consequently, the observed decline in Reflow's generation quality has been attributed to error accumulation without exploring the underlying mechanisms of collapse.

Simulation-free models and their variants—such as diffusion models \citep{song2019generative, song2020score, ho2020denoising}, flow matching \citep{lipman2022flow, pooladian2023multisample, tong2023improving}, and rectified flow 
% \zz{Rectified Flow or rectified flow?} 
\citep{liu2022flow}—have drawn increasing attention. Among these models, rectified flow stands out due to its rapid development and extensive foundational and large-scale work \citep{esser2024scaling}. Unlike conventional diffusion models, Rectified Flow employs a reflow procedure that iteratively uses self-generated data as training data to straighten the flow and improve sampling efficiency. While Rectified Flow can reduce the number of sampling steps, the quality of generated data tends to decline, particularly with a higher number of reflow steps. However, there is a lack of thorough analysis regarding this performance drop; prior studies on Rectified Flow have primarily focused on scaling up the model or applying distillation techniques \citep{lee2024improving, liu2023instaflow, esser2024scaling}. As a result, the observed decline in reflow’s generation quality is often attributed to error accumulation.

Rectified flow and model collapse share a similar setting—both involve retraining models with self-generated data—but deliver conflicting messages, as rectified flow produces more efficient models while model collapse highlights model failure. In this paper, we aim to resolve this conflict by studying the following questions: {\it Does rectified flow suffer from model collapse? If so, how can model collapse be prevented while leveraging the benefits of reflow?}
\vspace{-2mm}
\paragraph{Contribution} This paper investigates model collapse in diffusion models by (1) demonstrating that rectified flow also suffers from model collapse, (2) providing a theoretical analysis using a denoising autoencoder, and (3) proposing a novel method to prevent model collapse. Our contributions can be summarized as follows.
\vspace{-1mm}
\begin{itemize}[leftmargin=*]
\setlength\itemsep{-.3em}
\item \textbf{Empirical verification of model collapse.} Complementing existing works that empirically demonstrate diffusion/flow models suffer from model collapse in self-consuming training loops \citep{alemohammad2023selfconsuming}, we experimentally validate that the reflow training method also leads to a decline in model performance on both synthetic and real tasks. See \Cref{sec:related work} for the detailed description of related work.
% \zz{to weaken this point.}

\item \textbf {Theoretical analysis of model collapse.} We then delve into a theoretical analysis of MC. To facilitate the analysis, we study Denoising Autoencoders (DAEs) and uncover the underlying mechanisms that lead to performance degradation when DAEs are trained iteratively on their own outputs. To the best of our knowledge, we are the first to rigorously analyze model collapse in DAEs and, by extension, in diffusion models and Rectified Flow. Our analysis validates the causes of performance degradation arising from iterative training on self-generated data in Rectified Flow. In addition, we prove that incorporating real data can prevent model collapse during the recursive training of DAEs, supporting the recent development of using real data as an effective approach for mitigating model collapse \citep{bertrand2023stability}.  
\item \textbf{RA Reflow: Real-data Augmented Rectified flow} Following this analysis, a natural approach for preventing model collapse is by incorporating real data. {\it However, the absence of direct noise-image pairs poses a significant challenge in directly using real data with rectified flow.} Our proposed RA Reflow addresses this issue with a novel approach of incorporating real data by leveraging reverse processes. With balanced synthetic and real data, RA Reflow can straighten the flow trajectories effectively while maintaining training stability. We also propose an online RA Reflow that can greatly reduce the storage budget of naive reflow while progressively straightening the flow
% \zz{to say the advantages}.
We validate our methods through extensive experiments on standard image datasets. The results demonstrate that our approaches not only mitigate MC but also enhance sampling efficiency, allowing for high-quality image generation with fewer sampling steps. This confirms the effectiveness of our strategies in both theoretical and practical aspects.

\end{itemize}

\begin{table}[t]
\centering
\small % You can change to \scriptsize for even more compactness
\setlength{\tabcolsep}{3pt} % Adjust column spacing
\resizebox{\columnwidth}{!}{%
  \begin{tabular}{@{}llcc@{}} 
    \toprule
    \textbf{Methods} & \textbf{Variant} & \textbf{Eff. Sampling} & \textbf{Collapse Mitigating} \\ 
    \midrule
    \multirow{5}{*}{\makecell[c]{ \\ \\ \\ Diffusion/\\Flow Model}} 
      & \makecell{\textbf{RA Reflow} \\ \textbf{Ours}}      & \cmark & \cmark \\ 
      & \makecell{DDPM \\ \citep{ho2020denoising}}      & \xmark & \xmark \\ 
      & \makecell{FM \\ \citep{lipman2022flow}}       & \cmark (Weak) & \xmark \\ 
      & \makecell{OTCFM \\ \citep{tong2023improving}}    & \cmark (Weak) & \xmark \\ 
      & \makecell{RF \\ \citep{liu2022flow}}       & \cmark & \xmark \\ 
    \midrule
    Distillation & \makecell{CD/CT \\ \citep{song2023consistency}} & \cmark (1-step) & Unknown \\ 
    \midrule
    \multirow{4}{*}{\makecell[c]{ \\ \\ \\ Collapse\\Mitigating}} 
      & \makecell{MAD \\ \citep{alemohammad2023selfconsuming}}      & \xmark & \cmark \\ 
      & \makecell{Stability \\ \citep{bertrand2023stability}} & \xmark & \cmark \\ 
      & \makecell{MCI \\ \citep{gerstgrasser2024model}}      & N/A & \cmark \\ 
      & \makecell{MCD \\ \citep{dohmatob2024model}}      & N/A & \cmark \\ 
    \bottomrule
  \end{tabular}
}
\caption{Comparison of methods regarding efficient sampling and model collapse mitigating. \cmark\ and \xmark\ indicate feature presence or absence; "Weak" denotes limited capability, and "Unknown" or "N/A" means insufficient information or not applicable.}
\label{tab:methods_comparison}
\vspace{-2em}
\end{table}

% \begin{table*}[t]
% \centering
% \label{tab:methods_comparison}
% \footnotesize 
% \resizebox{0.9\textwidth}{!}{%
%   \begin{tabular}{@{}llcc@{}} 
%     \toprule
%     \multirow{2}{*}{\textbf{Methods}} & \multirow{2}{*}{\textbf{Variant}} & \multicolumn{2}{c}{\textbf{Performance}} \\
%     \cmidrule(lr){3-4}
%     & & \textbf{Efficient Sampling} & \textbf{Model Collapse Avoid} \\ 
%     \midrule
%     \multirow{5}{*}{Diffusion/Flow Model} 
%       & \textbf{Ours}      & \cmark & \cmark \\ 
%       & DDPM \citep{ho2020denoising}      & \xmark & \xmark \\ 
%       & FM \citep{lipman2022flow}       & \cmark\ (Weak) & \xmark \\ 
%       & OTCFM \citep{tong2023improving}    & \cmark\ (Weak) & \xmark \\ 
%       & RF \citep{liu2022flow}       & \cmark & \xmark \\ 
%     \midrule
%     Distillation & CD/CT \citep{song2023consistency} & \cmark\ (1-step) & Unknown \\ 
%     \midrule
%     \multirow{4}{*}{Collapse Avoid} 
%       & MAD \citep{alemohammad2023selfconsuming}      & \xmark & \cmark \\ 
%       & Stability \citep{bertrand2023stability} & \xmark & \cmark \\ 
%       & MCI \citep{gerstgrasser2024model}      & N/A & \cmark \\ 
%       & MCD \citep{dohmatob2024model}      & N/A & \cmark \\ 
%     \bottomrule
%   \end{tabular}
% }
% \caption{Comparison of various methods regarding efficient sampling and model collapse avoidance. Symbols \cmark\ and \xmark\ indicate the presence or absence of a feature, respectively; "Weak" denotes limited capability, and "Unknown" or "N/A" indicates insufficient information or not applicable.}
% \vspace{-1em}
% \end{table*}

\vspace{-5mm}
\section{Related Work}
\label{sec:related work}
\subsection{Model Collapse in Generative Models}

The generation of synthetic data by advanced models has raised concerns about MC, where models degrade when trained on their own outputs. Although large language models and diffusion models are primarily trained on human-generated data, the inadvertent inclusion of synthetic data can lead to self-consuming training loops \citep{alemohammad2023selfconsuming}, resulting in performance degradation \citep{shumailov2023curse}. Empirical evidence of MC has been observed across various settings \citep{Hataya_2023_ICCV, martínez2023combining, bohacek2023nepotistically}. Theoretical analyses attribute the collapse to factors like sampling bias and approximation errors \citep{shumailov2023curse, dohmatob2024model}. While mixing real and synthetic data can maintain performance \citep{bertrand2023stability}, existing studies often focus on maximum likelihood settings without directly explaining MC. Our work extends these analyses to simulation-free generative models like diffusion models and flow matching, specifically addressing MC in the Reflow method of Rectified Flow and proposing more efficient training of Rectified flow.

\subsection{Efficient Sampling in Generative Models}

Achieving efficient sampling without compromising quality is a key challenge in generative modeling. GANs \citep{goodfellow2014generative} and VAEs \citep{kingma2013auto} offer fast generation but face issues like instability and lower sample quality. Diffusion models \citep{song2020score} and continuous normalizing flows \citep{chen2018neural, lipman2022flow, albergo2022building},  produce high-fidelity outputs but require multiple iterative steps, slowing down sampling. To accelerate sampling, methods such as modifying the diffusion process \citep{song2020denoising, bao2021analytic, dockhorn2021score}, employing efficient ODE solvers \citep{ludpm, dockhorn2022genie, zhang2022fast}, and using distillation techniques \citep{salimans2022progressive} have been proposed. Consistency Models \citep{song2023consistency, kim2023consistency, yang2024consistency} aim for single-step sampling but struggle with complex distributions. Rectified Flow and its Reflow method \citep{liu2022flow, lee_improving_2024} promise efficient sampling by straightening flow trajectories, needing fewer steps. However, we will show that they are prone to MC due to training on self-generated data, and existing mitigating methods are ineffective due to the lack of noise-image pairs. Our work addresses this gap by proposing methods to prevent MC in Rectified Flow.

\begin{figure*}[t]
  \centering
  \includegraphics[width=0.85\linewidth]{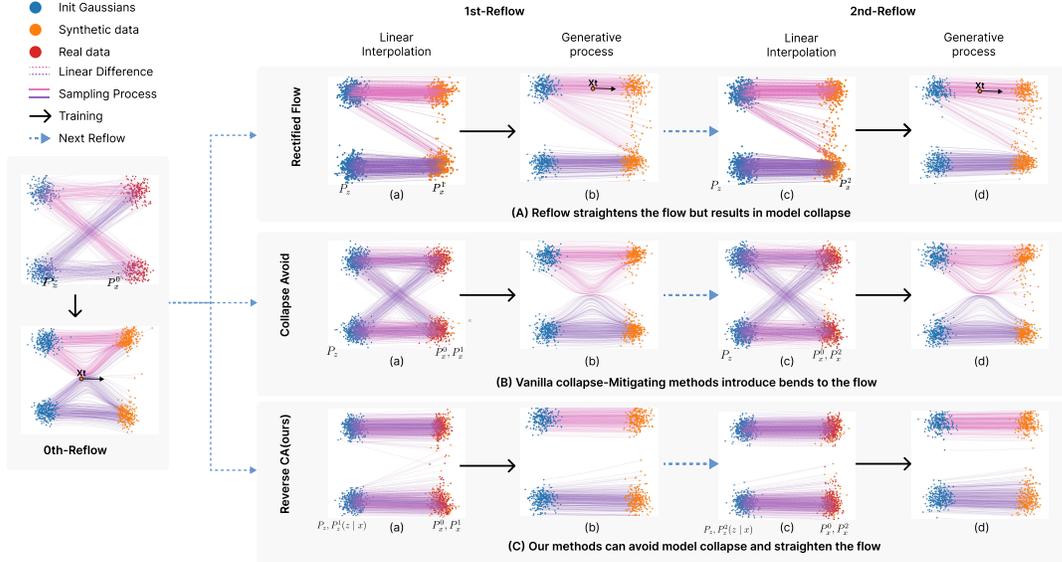}
  \caption{\textbf{2D multi-Gaussian experiment demonstration.} \textcolor{red}{(A)} Rectified Flow rewires trajectories to eliminate intersecting paths, transforming from \textcolor{blue}{$(a)$} to \textcolor{blue}{$(b)$}. We then take noise samples from the distribution $p^z$ and their corresponding generated samples from the synthetic distribution $p_1^x$ to construct noise-target sample pairs (\textcolor{blue}{blue} to \textcolor{orange}{orange}) and linearly interpolate them at point \textcolor{blue}{$(c)$}. In Reflow, Rectified Flow is applied again from \textcolor{blue}{$(c)$} to \textcolor{blue}{$(c)$} to straighten the flows. This procedure is repeated recursively. \textcolor{red}{(B)} Since iterative training on self-generated data can cause MC, we can incorporate real data (\textcolor{red}{shown in red}) during training to prevent collapse. \textcolor{red}{(C)} However, adding real data introduces additional bends to the Rectified Flow because the pairs of real data and initial Gaussian samples are not pre-paired. Our method employs reverse sampling generated real-noise pairs (\textcolor{red}{red} to \textcolor{blue}{blue}) to avoid MC while simultaneously straightening the flow.}
  \label{fig:2D gaussian methods compare}
    \vspace{-3mm}
\end{figure*}

\vspace{-3mm}
\section{Preliminaries}

% \subsection{Flow matching}\label{sec:flow-matching}
% In contrast to likelihood-based training, flow matching (FM) allows training CNFs simulation-free, i.e. without integrating the vector field and evaluating the Jacobian, making the training significantly faster. Hence, the flow matching enables scaling to larger models and larger systems with the same computational budget.
% The flow matching training objective \cite{lipman2022flow, liu2022flow} regresses   $v_{\theta}(t, x)$ to some target vector field $u_t(x)$ via
% \begin{equation}\label{eq:fm-loss}
%     \mathcal{L}_{\mathrm{FM}}(\theta)=\mathbb{E}_{t\sim[0,1],x \sim p_t(x)}\left|\left|v_{\theta}(t, x)- u_t(x)\right|\right|_2^2.   
% \end{equation}
% In practice, $p_t(x)$ and $u_t(x)$ are intractable,  but it has been shown in \cite{lipman2022flow} that the same gradients can be obtained by using the conditional flow matching loss:
% \begin{equation}\label{eq:conditional-fm-loss}
%     \mathcal{L}_{\mathrm{CFM}}(\theta)=\mathbb{E}_{t\sim[0,1], x\sim p_t(x|z)}\left|\left|v_{\theta}(t, x)- u_t(x|z)\right|\right|_2^2. 
% \end{equation}
% The vector field $u_t(x)$ and corresponding probability path $p_t(x)$ are given in terms of the conditional vector field $u_t(x|z)$ as well as the conditional probability path $p_t(x|z)$ as
% \begin{align}
%     p_t(x) = \int p_t(x|z) p(z) dz \quad \textrm{and}\quad
%     u_t(x) = \int \frac{p_t(x|z) u_t(x|z)}{p_t(x)} p(z) dz,
% \end{align}
% where $p(z)$ is some arbitrary conditioning distribution independent of $x$ and $t$.

\subsection{Flow Matching}\label{sec:flow-matching}

Flow Matching (FM) is a training paradigm for Continuous Normalizing Flow (CMF) \citep{chen2018neural} that enables simulation-free training, avoiding the need to integrate the vector field or evaluate the Jacobian, thereby significantly accelerating the training process \citep{lipman2022flow, liu2022flow, albergo2022building}. This efficiency allows scaling to larger models and systems within the same computational budget. Let $\mathbb{R}^d$ denote the data space with data points $\vx \in \mathbb{R}^d$. The goal of FM is to learn a vector field $v_\theta(t, \vx): [0,1] \times \mathbb{R}^d \rightarrow \mathbb{R}^d$ such that the solution of the following ODE transports noise samples $\vx_0 \sim p_0$ to data samples $\vx_1 \sim p_1$:
\begin{equation}\label{eq:ode}
        \frac{d \phi_\vx(t)}{dt} = v_\theta(t, \phi_\vx(t)), 
        \phi_\vx(0) = \vx.
\end{equation}
Here, $\phi_\vx(t)$ denotes the trajectory of the ODE starting from $\vx_0$. FM aims to match the learned vector field $v_\theta(t, \vx)$ to a target vector field $u_t(\vx)$ by minimizing the loss:
\begin{equation}\label{eq:fm-loss}
    \mathcal{L}_{\mathrm{FM}}(\theta) = \mathbb{E}_{t \sim [0,1], x \sim p_t(\vx)} \left\| v_\theta(t, \vx) - u_t(t, \vx) \right\|_2^2,
\end{equation}
% \zz{the expectation is over $\vx$ while the expression is about $\vx_t$. Same for the following two equations.}
where $p_t$ is the probability distribution at time $t$, and $u_t$ is the ground truth vector field generating the probability path $p_t$ under the marginal constraints $p_{t=0} = p_0$ and $p_{t=1} = p_1$. However, directly computing $u_t(\vx)$ and $p_t(\vx)$ is computationally intractable since they are governed by the continuity equation \citep{villani2009optimal}: $\partial_t p_t(\vx) = - \nabla \cdot \left( u_t(\vx)p_t(\vx) \right)$.

To address this challenge, \citet{lipman2022flow} proposes regressing $v_\theta(t, x)$ on a conditional vector field $u_t(x|z)$ and the conditional probability path $p_t(x|z)$, where $z \sim p(z)$ is an arbitrary conditioning variable independent of $x$ and $t$ (normally we set $p(z)$ as Gaussian Distribution). This gives
\begin{equation}\label{eq:cfm-loss}
\small
    \mathcal{L}^{\mathrm{CFM}}_\theta = \mathbb{E}_{t \sim [0,1], \vz \sim p(\vz), \vx \sim p_t(\vx|\vz)} \left\| v_\theta(t, \vx) - u_t(t, \vx|z) \right\|_2^2.
\end{equation}
Two objectives \eqref{eq:fm-loss} 
% \zz{could we change the style to (2)? not sure why eqref does not give this one here.} 
and \eqref{eq:cfm-loss}  share the same gradient with respect to $\theta$, while \eqref{eq:cfm-loss} can be efficiently estimated as long as the conditional pair $u_t(t,x|z), p_t(x|z)$ is tractable. By setting $x_t = tz +(1-t)x$, $u_t(t,x_t|z) = \frac{z - x}{1-t}$ we get the loss of Rectified flow \citep{liu2022flow}:
\begin{equation}\label{eq:RF-loss}
\small
    \mathcal{L}^{\mathrm{RF}}_\theta = \mathbb{E}_{t \sim [0,1], \vz \sim p(\vz), \vx_1 \sim p_1} \left\| v_\theta(t, t\vz +(1-t)\vx) - (\vx - \vz) \right\|_2^2.
\end{equation}
\vspace{-2mm}
\subsection{Rectified Flow and Reflow}\label{sec:rectified-flow}

Rectified Flow (RF) \citep{liu2022flow, liu2022rectified, liu2023instaflow} extends FM by straightening probability flow trajectories, enabling efficient sampling with fewer number of function evaluations (NFEs). In standard FM, the independent coupling \( p_{\vx\vz}(\vx, \vz) = p_{\vx}(\vx) p_{\vz}(\vz) \) 
% \zz{$\vx$ is not consistent with previous $x$} 
results in curved ODE trajectories, requiring a large NFEs for high-quality samples. RF addresses this by iteratively retraining on self-generated data to rewire and straighten trajectories.

The \textit{Reflow} algorithm \citep{liu2022flow} implements this idea by recursively refining the coupling between $x$ and $z$. Starting with the initial independent coupling $p^{(0)}_{\vx_0, \vz}(\vx_0, \vz) = p_{\vx_0}(\vx_0) p_{\vz}(\vz)$, we can train the first Rectified flow $\theta_0$ by RF-loss \eqref{eq:RF-loss} using stochastic interpolation data as input (see \Cref{fig:2D gaussian methods compare} 0th-Reflow). Then, we can generate noise-image pairs because we can draw $(\vx_1, \vz)$ following $dx_t = v_{\theta_0}(\vx_t, t) dt$ starting from $\vz\sim\mathcal{N}$ which means we can have $p^{(1)}_{\vx_1\ \vz}(\vx_1, \vz) = p_{\vx_1}(\vx_1) p_{\vz}(\vz)$ 
% \zz{shouldn't $\vx_1$ depend on $\vz$?} \zz{the notation $p_{\vx_1}(\vx)$ seems odd? $\vx_1$ already means a random variable, then why not $p_{\vx_1}(\vx_1)$? or $p_{1}(\vx)$?  } 
to start the reflow. Reflow generates an improved coupling $p^{(k+1)}_{\vx_{k+1}\vz}(\vx, \vz)$ 
% \zz{the notation is not consistent with previous using $p^{(k)}_{\vx_{k}\vz}(\vx, \vz)$} 
at each iteration $k$ by:
\begin{enumerate}[noitemsep,topsep=0pt]
    \item Generating synthetic pairs $(\vx_k, \vz)$ sampled from the current coupling $p^{(k)}_{\vx_{k}\vz}(\vx_k, \vz)$.
    \item Training a new rectified flow $\theta_{k+1}$ by \eqref{eq:RF-loss} using these synthetic pairs.
\end{enumerate}
We denote the vector field resulting from the $k$-th iteration as the \textit{$k$-Reflow}. This process aims to produce straighter trajectories, thus reducing the NFEs required during sampling. However, existing literature on learning with synthetic data \citep{alemohammad2023selfconsuming,briesch2023large}, though not specifically focusing on rectified flow, suggests that iterative training on self-generated data can lead to \emph{model collapse}, where performance degrades over iterations. Furthermore, existing methods for avoiding MC are ineffective for rectified flow because incorporating real data does not provide the necessary noise-image pairs required for Reflow training. In the next two sections, we analyze MC in diffusion models and rectified flow and introduce a novel method for its prevention.

\begin{figure*}[!ht]
  \centering
  \includegraphics[width=0.85\linewidth]{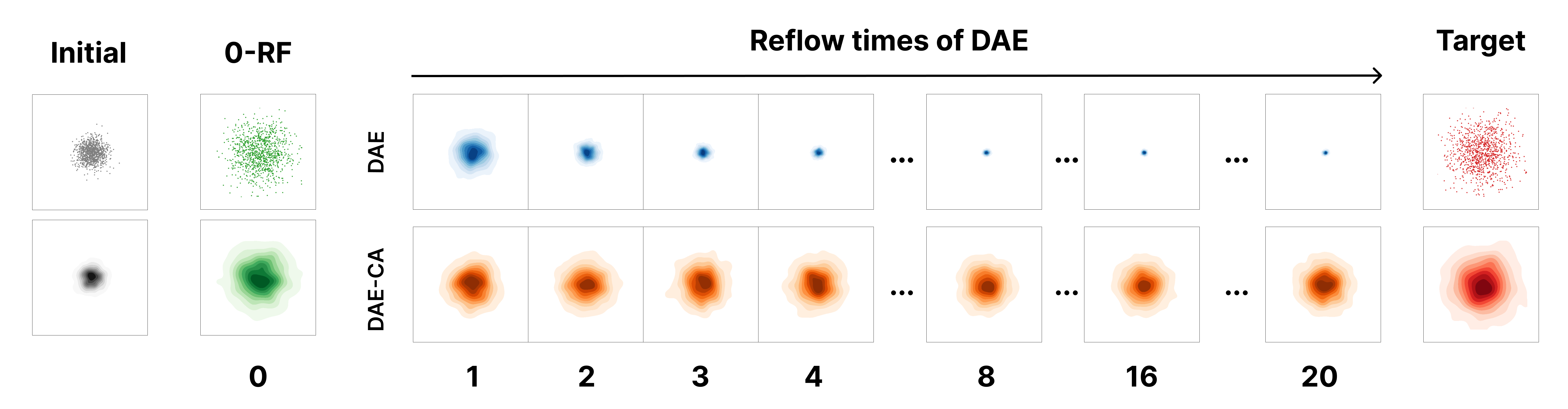}
  \caption{Reflow Process of the DAE on a 4-D Gaussian Distribution. The figure visualizes a slice of the distribution along dimensions 0 and 1. Both kernel density estimation plots and sample points are shown for the initial and target distributions. }
  \label{fig: DAE_KDE}
  \vspace{-5mm}
\end{figure*}

\vspace{-3mm}
\section{Theoretical Analysis of Model Collapse }
% \subsection{Connection between Denoising Autoencoders and Diffusion Models}
Having introduced diffusion/flow models and Reflow method in the previous section, we now turn to Denoising Autoencoders (DAEs), which are theoretically linked to diffusion models via score matching. Here, we show how relying solely on model-generated samples in Reflow of DAEs can lead to collapse, illustrate how incorporating real data prevents this degeneracy, and connect these insights to the Rectified Flow framework.

% \zz{The jump to DAE is abrupt. We can give some discussion here, and also provide a summary of this section.}
\subsection{Denoising Autoencoders and Diffusion Models}
\label{sec:dae-diffusion}

DAEs are closely related to diffusion models through the concept of score matching \citep{song2019generative, song2020score}. Under certain conditions, training a DAE implicitly performs score matching by estimating the gradient of the log-density of the data distribution \citep{vincent2011connection}. Specifically, given data $\mathbf{\vx}$ and Gaussian noise $\mathbf{\epsilon} \sim \mathcal{N}(0, \sigma^2 \mathbf{I})$, the DAE minimizes the reconstruction loss: 
% \zz{here $\mathbf{x}$ is not consistent with previous $\vx$ or $x$. Need to unify all of them.}
% \textcolor{blue}{
\begin{equation}\label{eq:DAE_training} 
\small
\mathcal{L}_{\mathrm{DAE}} = \mathbb{E}_{\mathbf{x}, \mathbf{\epsilon}} \left\| f_\theta(\mathbf{x} + \mathbf{\epsilon}) - \mathbf{x} \right\|^2, 
\end{equation} 
% }
where $f_\theta$ is the DAE parameterized by $\theta$. The residual between the output and input approximates the scaled score function: \begin{equation}\label{eq
} f_\theta(\mathbf{x} + \mathbf{\epsilon}) - \mathbf{x} \approx \sigma^2 \nabla_{\mathbf{x}} \log p(\mathbf{x} + \mathbf{\epsilon}). 
\end{equation}
We demonstrate that the training objectives of diffusion models and Flow Matching methods, such as Rectified Flow, can be unified, differing only in parameter settings and affine transformations \citep{esser2024scaling}. Specifically, diffusion models are special cases of Continuous Normalizing Flow trajectories \citep{lipman2022flow, liu2022flow}. Consequently, analyzing MC in DAEs is essential for understanding MC in diffusion models and Rectified Flow. Since DAEs learn to denoise and approximate the score function, examining their behavior under iterative training on self-generated data can reveal degradation mechanisms in more complex generative models. In this work, we focus on a simplified scenario where a DAE is recursively trained on its own generated data, enabling an analytical study of MC.

\subsection{Analysis of DAE  with Recursively Learning}
% \zz{to briefly add some motivation for studying this setting.} \mh{4.1 is part of the description, I will continue here}
To better understand the mechanisms behind MC, we investigate a simplified scenario where a linear DAE is trained recursively on the data it generates. Studying this setting provides valuable insights into how errors can accumulate over iterations, leading to performance degradation, which is challenging to analyze in more complex models. 

Following the settings similar to \citet{pretorius2018learning}, we consider a two-layer neural network denoted by $f_\theta(\vx): \mathcal{X} \to \mathcal{X}$, which can be expressed in matrix form as $f_\theta(\vx) = \mW_2\mW_1 \vx$, where $\mW_2\in \mathbb{R}^{d\times d'}$, $\mW_1\in \mathbb{R}^{d' \times d}$ represents the weights of the network. We aim to optimize the following training objectives:
\begin{equation}\label{linear DAE Loss}
    \min_{\theta} \mathcal{L}(\theta) = \min_{\theta} \mathbb{E}_{\tilde{\vx} \sim p(\vx \mid \vz),  \vz \sim \mathcal{N}} \left[\left\|f_\theta(\tilde{\vx}) - \vx\right\|^2_2\right],
\end{equation}
where \( \vz \sim \mathcal{N}(0, \sigma^2) \) denotes Gaussian noise, \( \vx \sim p_1 \) represents the original training data, and \( \tilde{\vx} \) is a perturbed version of \( \vx \), defined by \( \tilde{\vx} = \alpha \vx + \beta \vz \). The parameters $\alpha$ and $\beta$ are affine transformations that depend on the variable $t$. Here, we set \( \alpha = \beta = 1 \) for the simplicity of analysis. In practice, given a finite number of training samples, $\mX=\begin{bmatrix} \vx_1 & \cdots & \vx_n\end{bmatrix}$, we learn the DAE by solving the following empirical training objectives
\begin{equation}\label{eq:empirical-DAE-Loss}
  \theta^\star(\mX):= \argmin_{\theta} \sum_i\mathbb{E}_{\vz \sim \mathcal{N}(0,\sigma^2 I)} \left[\left\|f_\theta(\vx_i + \vz) - \vx_i\right\|^2_2\right],
\end{equation}
where $\theta^\star(\mX)$  emphasizes the dependence of the solution on the training samples $\mX$. 

\paragraph{Self-consuming recursively training} \label{Setting: Synthetic Data Generation Process} 
Now we formulate the Reflow of linear DAE. Suppose we have training data $\mX = \begin{bmatrix} \vx_1 & \cdots & \vx_n\end{bmatrix}$ with $\vx_i = \mU^\star \mU^{\star\top} \va_i, \va_i\sim \calN(0,I)$. 
\begin{definition}[Self-consuming training loops for DAE]
\label{meth: Reflow of DAE}
Start with $\mX_1 = \mX$, in the $j$-th iteration with $j \ge 1$, the scheme for generating synthetic data is outlined as follows.
\begin{enumerate}[noitemsep,topsep=0pt]
\item Fit DAE: $(\mW_2^j,\mW_1^j) = \theta^\star(\mX_j)$ by solving \eqref{eq:empirical-DAE-Loss} with training data $\mX_j$
\item Generate synthetic data for the next iteration: $\mX_{j+1} = \mW_2^{j}\mW_1^j(\mX_j + \mE_j)$, where each column of the noise matrix $\mE_j$ is iid sampled from $\calN(0,\hat \sigma^2/n^2 I)$.
\end{enumerate} 
\end{definition} 

% \phantomsection
% \label{meth: Reflow of DAE}
% \begin{enumerate}
% \item Fit DAE: $(\mW_2^j,\mW_1^j) = \theta^\star(\mX_j)$ by solving \eqref{eq:empirical-DAE-Loss} with training data $\mX_j$
% \item Generate synthetic data for the next iteration: $\mX_{j+1} = \mW_2^{j}\mW_1^j(\mX_j + \mE_j)$, where each column of the noise matrix $\mE_j$ is iid sampled from $\calN(0,\hat \sigma^2/n^2 I)$.
% \end{enumerate} 

The following result establishes the occurrence of model collapse during the recursive training of DAEs.  
\begin{tcolorbox}[colback=gray!10, colframe=white, sharp corners, boxrule=0pt]\small
\begin{theorem} \label{prop: DAE collapse} In the self-consuming recursively training process \Cref{meth: Reflow of DAE}, suppose that the variance of the added noise is not too large, i.e., $\hat \sigma \le C \sigma$ for some universal constant $C$. Then, with probability at least $1-2je^{-n}$, the learned DAE suffers from MC as
    \begin{equation}
     \|\mW_2^j \mW_1^j\|^2 \le \frac{\|\mX\|^2}{\sigma^2} (\frac{\|\mX\|^2}{\|\mX\|^2 + \sigma^2})^{j-1}.
    \end{equation}
\end{theorem}
\end{tcolorbox}
% Theorem~\ref{prop: DAE collapse} shows that weight matrices of DAE decrease geometrically across recursively training steps, iterations, implying that the network’s output is pushed closer to zero whenever it relies solely on its own previously generated samples for training. This repeated self-consuming training drives the network toward a degenerate solution, illustrating why MC emerges. 
%The proof of Theorem~\ref{prop: DAE collapse} is given in Appendix~\ref{Proofs and Formulations}. 
Theorem~\ref{prop: DAE collapse} shows that weight matrices of DAE decrease geometrically across recursively training steps, iterations, implying that repeated self-consuming training drives the network toward a degenerate solution. Our proof of Theorem~\ref{prop: DAE collapse} in Appendix~\ref{Proofs and Formulations} illustrates that the implicit regularization of noise in the DAE \citep{pretorius2018learning} induces a distribution shift in the generated synthetic data, contributing to the emergence of MC. While MC has been empirically observed in diffusion models \citep{alemohammad2023selfconsuming,bertrand2023stability}, our result offers a clearer picture in a linear DAE setting, providing a theoretical rate of collapse and identifying the critical role of introducing new noise or real data to avoid degeneracy.
% \zz{A missing part is to provide a high-level discussion why MC happens from our analysis, and how it either aligns with existing analysis, or provides new perspectives.}
\begin{remark}[Connection to Diffusion Models]\label{rem
} The primary gap between diffusion models and a sequence of end-to-end DAEs lies in the initial step of the diffusion process. This perspective aligns with discussions in \citet{Zhang_2024_CVPR}, which examine the gap in the first step of diffusion models. For a detailed explanation, see Appendix~\ref{remark: DAE and diffusion model}. \end{remark}
\paragraph{Does Model Collapse Occur in Rectified Flow?}
Building on our analysis of MC in DAEs, we investigate whether a similar collapse occurs in Rectified Flow. Despite the differences between DAEs and Rectified Flow, we hypothesize that MC can still manifest in Rectified Flow when trained iteratively on self-generated data.
\begin{tcolorbox}[colback=gray!10, colframe=white, sharp corners, boxrule=0pt] \small
\begin{proposition}
\label{prop:MCRF} 
Let ${v_{\theta_j}(t, \vx)}_{j=1}^\infty$ be a sequence of vector fields trained via Reflow in Rectified Flow. As $j \to \infty$, due to the sampling process of Rectified Flow, the generated result $\vx_{j,1}$ at time $t=1$ (i.e., the output of the $j$-th Reflow iteration) converges to a constant vector, indicating model collapse. \end{proposition} 
\end{tcolorbox}
To test this hypothesis, we conducted experiments with Rectified Flow under iterative training. Our empirical results indicate that, without incorporating real data, the performance of Rectified Flow degrades over successive Reflow iterations, consistent with MC. For a detailed theoretical analysis and proof supporting this hypothesis, please refer to Appendix~\ref{prof: Model Collapse in Rectified Flow}.
\subsection{Preventing MC by Incorporating Real Data}
% \textcolor{blue}{
Incorporating real data into the training process is a strategy to prevent MC in generative models \citep{bertrand2023stability, alemohammad2023selfconsuming, gerstgrasser2024model}. Mixing real and synthetic data helps maintain performance and prevents degeneration caused by over-reliance on self-generated data. However, this performance has been empirically verified with limited theoretical analysis; the existing analysis is primarily focused on regression problem settings \cite{dohmatob2024model, fu2024towards, gerstgrasser2024model, dohmatob2024strong}.  
% \zz{However, this performance has been empirically verified with limited theoretical analysis; the existing analysis is primarily focused on regression problems. @Minhao, is this correct? Help add references. We need to mention the limitations of existing work.} 
Inspired by these approaches, we extend the analysis of DEA by integrating real data. Recall the settings in \ref{Setting: Synthetic Data Generation Process}, we modify the synthetic data generation scheme by adding real data. Specifically, we augment the current synthetic data with real data by setting \(\hat{\mX}_j = \begin{bmatrix} \mX_j & \mX \end{bmatrix}\) and solving $(\mW_2^j,\mW_1^j) = \theta^\star(\hat\mX_j)$ in step 1 of \Cref{meth: Reflow of DAE}. To analyze the impact of adding real data, we present the following proposition (detailed settings and proof see Appendix~\ref{meth: Reflow of DAE with real data}):
% }
\begin{tcolorbox}[colback=gray!10, colframe=white, sharp corners, boxrule=0pt] \small
\begin{proposition} \label{prop: DAE not collapse}
In the self-consuming recursively training process \Cref{meth: Reflow of DAE} with adding real data, suppose that the variance of the added noise is not too large, i.e., $\hat \sigma \le C \sigma$ for some universal constant $C$. Then, with probability at least $1-2je^{-n}$,  the learned DAE does not suffer from model collapse as
\begin{equation}
\|\mW_2^j \mW_1^j\|^2 \geq \frac{\|\mX\|^2}{2\|\mX\|^2 + \sigma^2}.
\end{equation}
\end{proposition}
\end{tcolorbox}
% \textcolor{blue}{
Compared to Theorem 1, Proposition \ref{prop: DAE not collapse} 
% \zz{ref to the wrong place}
shows that by incorporating real data into the synthetic data generation process, the learned DAE mitigate model collapse, maintaining a fixed lower bound on the weight norm. In contrast, Theorem~\ref{prop: DAE collapse} indicates that without adding real data, the DAE's weight norm decreases exponentially with the number of iterations, leading to model collapse.

\vspace{-2mm}
\section{Mitigating Model Collapse in Reflow}
Building upon our exploration of MC in simulation-free generative models, this section addresses this challenge within the Rectified Flow framework. Although Rectified Flow and its Reflow algorithm \citep{liu2022flow} achieve efficient sampling by straightening probability flow trajectories, they are susceptible to MC due to iterative training on self-generated data (see Figure~\ref{fig:2D gaussian methods compare}(A)). Our analysis, consistent with \citet{bertrand2023stability, gerstgrasser2024model}, shows that incorporating real data can mitigate collapse. However, integrating real data in Rectified Flow is challenging because it requires noise-image pairs that are not readily available, and directly pairing real images with random noise invalidates the Reflow training (see Figure~\ref{fig:2D gaussian methods compare}(B)).

To overcome this limitation, we generate the necessary noise-image pairs using the reverse ODE process, commonly used in image editing tasks \citep{wallace2023edict, zhang2023exact}. This allows us to obtain exact inverse image-noise pairs given the pre-trained model and real images. However, another challenge arises due to the insufficient number of real image-noise pairs; for example, CIFAR-10 provides only 50,000 real images, while Reflow requires over 5 million data pairs per iteration \citep{liu2022flow}. Our Gaussian experiments suggest that a synthetic-to-real data ratio of at least 7:3 is needed to effectively avoid collapse  (see Figure~\ref{fig:DAE_linear_curve}). Using the reverse SDE process with significant randomness \citep{meng2021sdedit} can increase the number of the pairs but leads to image-noise pairs dominated by randomness, undermining the purpose of straightening the flow (like the vanilla collapse-avoid methods Figure~\ref{fig:2D gaussian methods compare}(B)). 

Therefore, the question arises: \textit{How can we generate sufficient real image-noise pairs while maintaining forward-backward consistency?} %In the following sections, we detail the implementation of RA Reflow, which effectively mitigates MC while preserving the efficiency benefits of Rectified Flow.
To address this challenge, we propose \textbf{Real-data Augmented Reflow (RA Reflow)}, which can effectively mitigate MC while preserving the efficiency benefits of Rectified Flow.

\vspace{-2mm}
\subsection{Real-data Augmented Reflow (RA Reflow)}
With a trained vector field $v_{\theta}$, to capture both forward and backward mappings between noise and images, we build our training data using two types of image-noise pairs: 
\begin{enumerate}[noitemsep,topsep=0pt]
    \item Synthetic pairs $\{(\vz^{(i)}, \hat{\vx}^{(i)})\}$ obtained by sampling noise $\vz^{(i)} \sim \mathcal{N}(\mathbf{0},\mathbf{I})$ and propagating it forward through the ODE solver:
    \begin{equation}
    \hat{\vx}^{(i)} = \mathrm{ODE}_{v_{\theta}}(0, 1, \vz^{(i)}),
    \end{equation}
    \item Real reverse pairs $\{(\hat{\vz}^{(i)}, \vx^{(i)})\}$ obtained by taking real image $\vx^{(i)}$ and propagating it backward to noise:
    \begin{equation}
    \hat{\vz}^{(i)} = \mathrm{ODE}_{v_{\theta}}(1, 0, \vx^{(i)}).
    \end{equation}
\end{enumerate}
We then select $\lambda n$ synthetic pairs and $(1-\lambda)n$ real reverse pairs (where $\lambda\in[0,1]$ controls the ratio of synthetical and real images) to form the combined dataset:
\begin{equation}
\mathcal{D}_j 
= \left\{(\vz^{(i)}, \hat{\vx}^{(i)})\right\}_{i=1}^{\lambda n} 
\,\cup\, \left\{(\hat{\vz}^{(i)}, \vx^{(i)})\right\}_{i=1}^{(1 - \lambda)n}.
\end{equation}
This mixture preserves diversity while leveraging both generated and real samples. We use a first-order Explicit Euler method to implement the ODE solver 
\(
\mathrm{ODE}_{v_{\theta}}(t_0, t_1, \vx).
\)
To prevent overfitting to outdated real reverse pairs, we regenerate them every $\alpha$ training epochs. As space is limited, the full training procedure for RA Reflow is provided in Algorithm~\ref{algo:RCA}.

\vspace{-2mm}
\subsection{Improved Techniques for RA Reflow}
\label{sec:ora_reflow}

Although RA Reflow can effectively prevent MC and straighten the flow (Figure~\ref{fig:2D gaussian methods compare}\,(C)), it has high storage requirements. For instance, \citet{lee2024improving} report using over 40 GB of memory for ImageNet $64 \times 64$ just to store $\hat{\vx}^{(i)}$ in one Reflow iteration. To address this limitation, we consider the case where the regeneration parameter $\alpha \to 0 \footnote{$\alpha$ controls how often real-data pairs are refreshed. Setting $\alpha \to 0$ effectively means regenerating them after each mini-batch. In practice, $\alpha$ cannot literally be zero, and we typically choose a batch size of 128 or 256.}$,
% \zz{you mean $\alpha \to 1$? if it reflects the difference between epoch and mini-batch, maybe add a footnote to explain it} 
so that we generate synthetic noise-image pairs and real reverse image-noise pairs in each mini-batch without storing intermediate results. We call this \textbf{Online Real-Data Augmented Reflow (ORA Reflow)}. This strategy resembles the consistency distillation method of \citet{kim2023consistency} but differs in two key aspects: 
(1)~we do not rely on a fixed, pretrained teacher model, instead repeatedly straightening the flow over multiple iterations, and 
(2)~we do not assume the network can recover any arbitrary point on the generative path from any input. Instead, we maintain a straight flow that is both interpretable and well-understood as a progressive approximation of the optimal transport map~\citep{liu2022rectified}. Detailed steps are provided in Appendix~\ref{algo: OCAR-datail}.

Moreover, relying solely on a deterministic ODE means that the only randomness in training comes from $\vz\sim \mathcal{N}(\mathbf{0}, \mathbf{I})$. This limited variability may reduce the diversity of generated samples~\citep{zhang2023emergence}. To address this, we introduce a small amount of noise in the reverse pass via an SDE with scale $\sigma$ (e.g., $\sigma=0.001$), implemented using the Euler-Maruyama scheme for up to 100 steps. We refer to this method as \textbf{Real-data Argument Stochastic Reflow (RAS Reflow)}. This controlled noise injection boosts sample diversity without significantly disrupting flow straightening. Further implementation details can be found in Appendix~\ref{meth:S}.

\vspace{-3mm}
\section{Experiments}
\begin{figure}[t]
  \centering
  % \begin{subfigure}[b]{0.32\linewidth}
  %   \centering
  %   \includegraphics[width=\linewidth]{fig/paramters/linear_plot_20_rank.png}
  %   \caption{Rank of W}
  %   \label{fig:rank1}
  % \end{subfigure}
  % \hfill
  \begin{subfigure}[b]{0.49\linewidth}
    \centering
    \includegraphics[width=\linewidth]{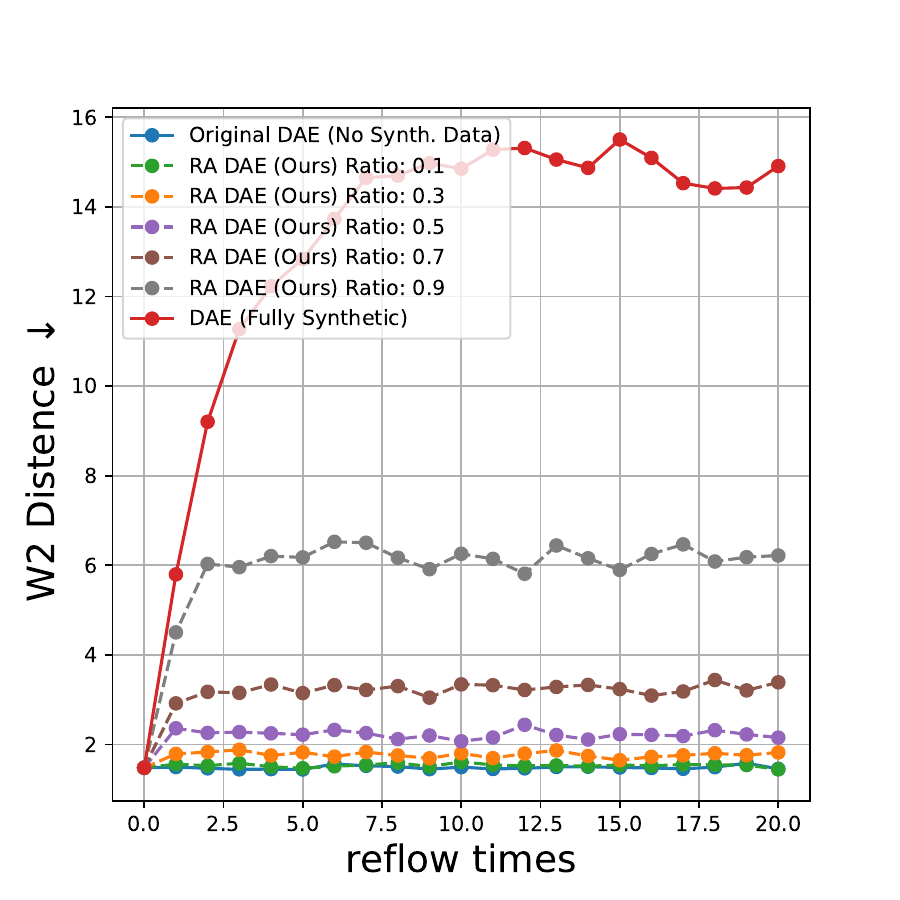}
    \caption{Wasserstain-2 Distance}
    % \label{fig:w2}
  \end{subfigure}
  \hfill
  \begin{subfigure}[b]{0.49\linewidth}
    \centering
    \includegraphics[width=\linewidth]{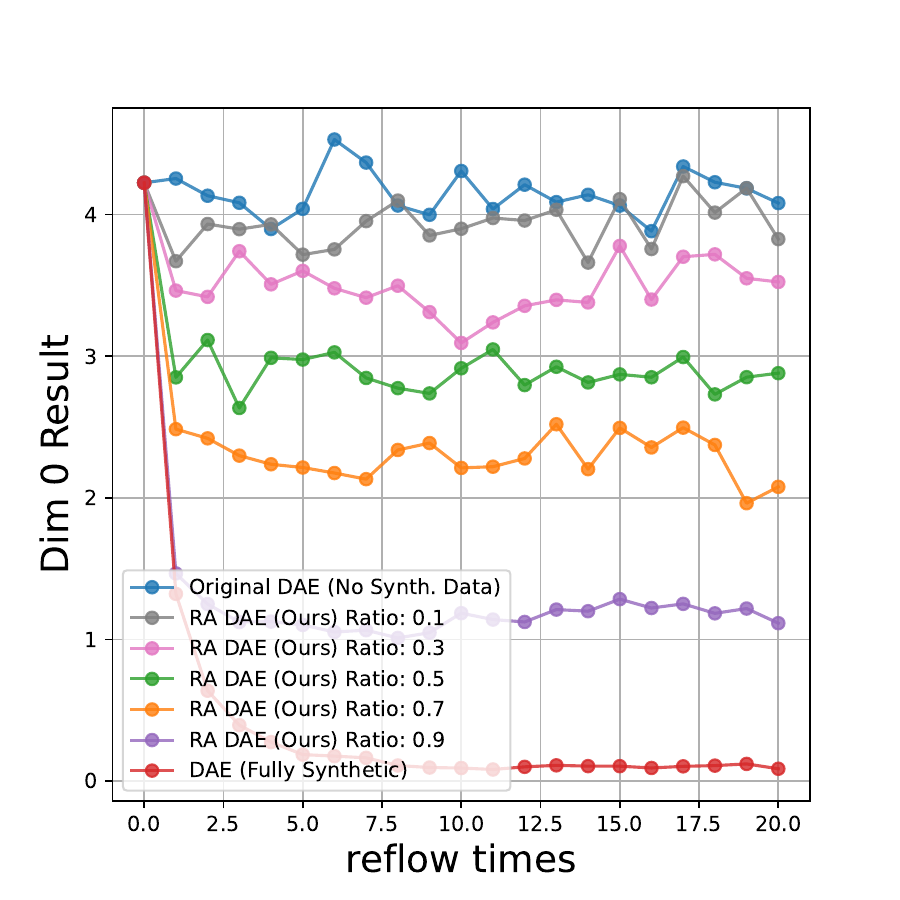}
    \caption{Dim 0}
    \label{fig:dim0}
  \end{subfigure}
  % \vspace{-3mm}
  \caption{
    Reflow experiment with DAE on 4D Gaussian. 
}
\vspace{-8mm}
\label{fig:DAE_linear_curve}
\end{figure}
% \begin{figure*}[t]
%   \centering
%   \begin{subfigure}[b]{0.32\linewidth}
%     \centering
%     \includegraphics[width=\linewidth]{fig/paramters/linear_plot_20_rank.png}
%     \caption{Rank of W}
%     \label{fig:rank1}
%   \end{subfigure}
%   \hfill
%   \begin{subfigure}[b]{0.32\linewidth}
%     \centering
%     \includegraphics[width=\linewidth]{fig/paramters/linear_plot_20_W2.png}
%     \caption{Wasserstain-2 Distance}
%     % \label{fig:w2}
%   \end{subfigure}
%   \hfill
%   \begin{subfigure}[b]{0.32\linewidth}
%     \centering
%     \includegraphics[width=\linewidth]{fig/paramters/linear_plot_20_dim0.png}
%     \caption{Dim 0}
%     \label{fig:dim0}
%   \end{subfigure}
%   % \vspace{-3mm}
%   \caption{
%     Results from the reflow experiment with DAE on 4D Gaussian. 
% }
% % \vspace{-7mm}
% \label{fig:DAE_linear_curve}
% \end{figure*}

In this section, we first validate our analysis of MC in DAEs and its extension to diffusion models and Rectified Flow. We then demonstrate that our proposed methods—RA Reflow, ORA Reflow, and ORAS Reflow—are capable of producing high-quality image samples on several commonly used image datasets. Additionally, we show that our methods provide a more efficient straightening of the sampling path, allowing for fewer sampling steps on CIFAR-10 \citep{krizhevsky2009learning}. Moreover, we demonstrate high-quality image generation on high-resolution datasets such as CelebA-HQ 256 \citep{karras2017progressive}, combined with latent space methods \citep{rombach2022high} commonly used in Rectified Flow \citep{dao2023flow, esser2024scaling}. We compare results using the Wasserstein-2 distance (W2, \citep{villani2009optimal}, lower is better), Fréchet Inception Distance (FID, \citep{heusel2017gans}, lower is better), and the number of function evaluations (NFE, lower is better). Due to limited space, we place the further settings of each experiment in the appendix.
\vspace{-6mm}
\subsection{Gaussian Task}
The intermediate columns of Figure~\ref{fig: DAE_KDE} illustrate the progression of the DAE Reflow process at different stages. They demonstrate that the original DAE Reflow leads to MC, whereas our proposed collapse-avoiding DAE Reflow maintains the integrity of the generated data.

Figure~\ref{fig:DAE_linear_curve} presents the key results from our DAE Reflow experiment on the 4D Gaussian task. The findings demonstrate that adding real data effectively prevents MC and maintains the integrity of the generated data. Specifically, incorporating real data helps maintain the rank of the weight matrix $\mW$ across Reflow iterations, our collapse-avoiding method consistently achieves a lower Wasserstein-2 distance compared to the original DAE Reflow, and the stability of the first principal component in PCA shows that our method effectively preserves the data structure over iterations. More details can be fund in Appenxix~\ref{app:Gaussian}
% shows the results from the reflow experiment with DAE on 4D Gaussian.(a) shows the rank of the weight matrix \( W \) across different reflow times. Here we set the testing bar as $2e-1$ which means we do SVD decomposition to $\mW_1$, we count the number of diagonal elements $geq 2e-1$ as the rank of weight matrix \( W \).  We show that adding real data can obviously avoid model collapse in terms of $\mW$-rank while the rank quickly goes to $0$ when fully using self-generated synthetic data. 
% (b) shows the performance in the metric of W2 distance between true target data and generated data distributions over reflow times. 
% (c) shows the results for the first principal component (Dim 0) of the data as reflow times increase. We compare the original DAE with no synthetic data and our DAE-CA model using various ratios of synthetic data (ranging from 0.1 to 0.9), as well as a fully synthetic DAE.

% \textbf{Set up for Rectified Flow}
% In the Reflow of the linear NN Rectified Flow verification experiment, we add one dimension to the $\mW_1$ in the context of time, so $\mW_1\mW_2:\sR^{d+1}\to\sR^d$ is the neural network we use. We also test the nonlinear neural network with three linear layers, SELU activation function, and an extra dimension for then first linear layer. The result is shown in Figure~\ref{fig:2D gaussian methods compare}

% \mh{To be continued, wait for images in osu serve}
\vspace{-2mm}
\subsection{Straight Flow and Fewer-Step Image Generation}

\begin{figure}[t]
    \centering
    \includegraphics[width=0.85\linewidth]{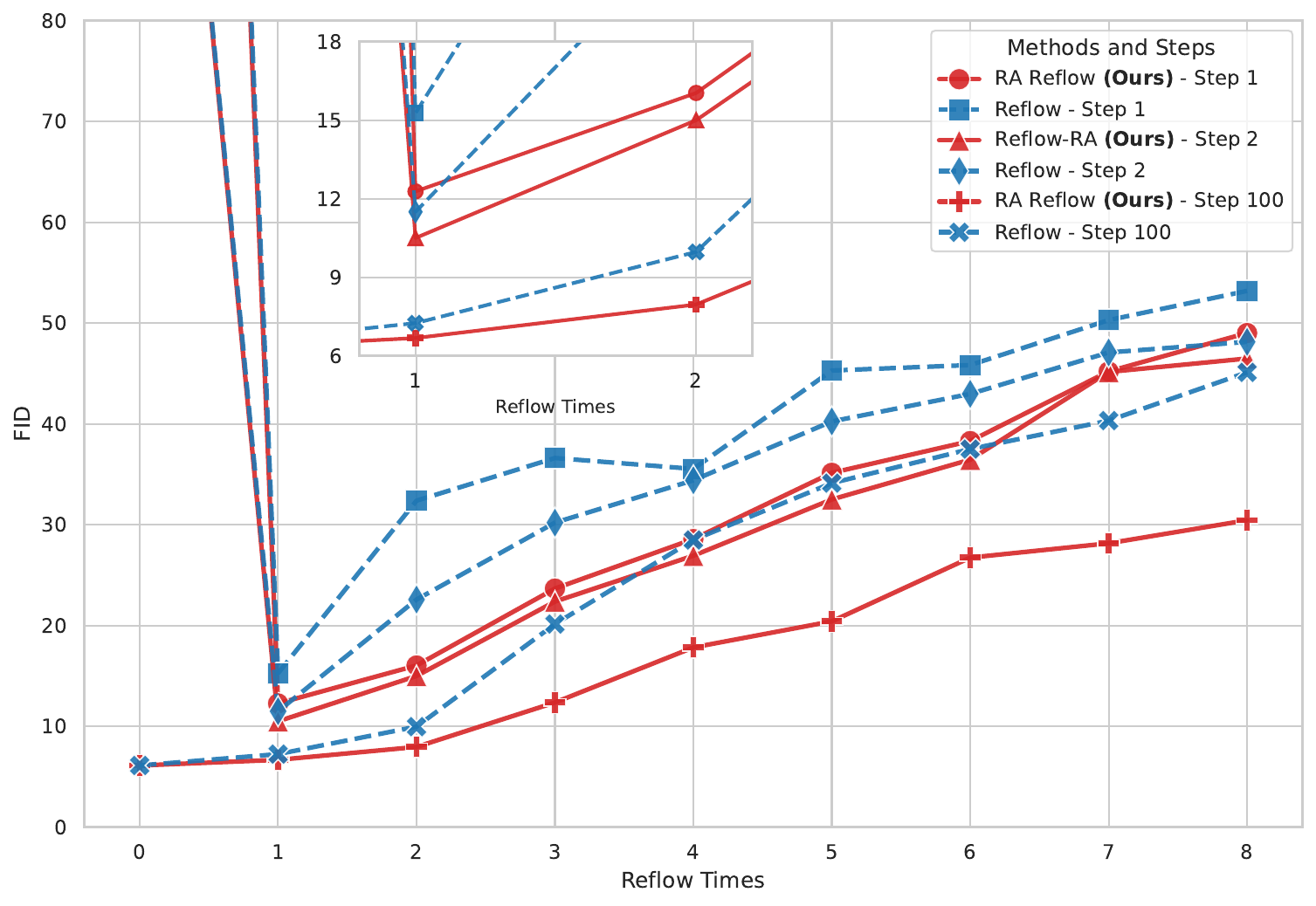}
    \vspace{-1mm}
    \caption{\textbf{Comparison of Reflow and RA Reflow.} Comparison of different methods. 
    We set $\lambda = 0.5, \alpha = 8$, and use a half-scale U-Net for the experiment. Full samples for Reflow processing are provided in Appendix~\ref{fig: 8-Reflow cifar10}.}
    \vspace{-8.9mm}
    \label{fig:fid_steps_comparison}
\end{figure}

% \begin{figure}[t]
%     \centering
%     % Adjust width as needed for layout
%     \includegraphics[width=0.9\linewidth]{fig/exper/smal_compare.pdf}
%     \caption{\textbf{Unselected generated image on CIFAR10} The setting is the same as Fig~\ref{fig:fid_steps_comparison}
%     % \zz{need to update the name of our reflow in the figure}
%     }
%     \label{fig:small_collapse_process}
% \end{figure}
% \begin{figure*}[t]
%   \centering
%   \begin{subfigure}[b]{0.47\linewidth}
%     \centering
%     \includegraphics[width=\linewidth]{fig/lowstep.png}
%     \caption{Comparison of methods}
%     \label{fig:Comparison_of_methods}
%   \end{subfigure}
%   \hfill
%   \begin{subfigure}[b]{0.47\linewidth}
%     \centering
%     \includegraphics[width=\linewidth]{fig/small_compare.pdf}
%     \caption{Samples demonstrating model collapse}
%     \label{fig:process}
%   \end{subfigure}
%   % \vspace{-2mm}
%   \caption{\textbf{Comparison of Reflow and RA Reflow} We set $\lambda = 0.5, \alpha = 8$ and use a half-scale U-Net for the experiment. See Appendix~\ref{fig: 8-Reflow cifar10} for full samples for reflow processing  }
%   \label{fig:fid_steps_comparison}
%   % \vspace{-3mm}
% \end{figure*}

\vspace{-1mm}
\textbf{RA Reflow}
% \textcolor{blue}{
Our experiments on CIFAR-10 show that Reflow achieves more efficient flows, enabling the use of fewer sampling steps. As illustrated in Figure~\ref{fig:fid_steps_comparison}, we observe the following key findings: \textbf{First}, 0-Reflow (vanilla Rectified Flow or FM) fails to enable 1- or 2-step sampling, whereas 1-RF and larger variants of RF can; \textbf{Second}, our RA Reflow method effectively prevents MC, resulting in more efficient training, as shown in Figure~\ref{fig: 8-Reflow cifar10}. Specifically, Table~\ref{tab:cifar10} and Table~\ref{tab:celeba} \footnote{ Best NFE is shown as FID/NFE using the DOPRI5 solver.} demonstrate that Rectified Flow trained with RA Reflow generates high-quality images using only a few sampling steps, underscoring the improvement in flow straightness. Detailed experimental settings and additional ablation studies can be found in Appendix~\ref{app:RCA_setup}.
% }

% Our experiments on CIFAR-10 demonstrate that Reflow achieves straighter flows, allowing for lower sampling steps. As shown in Figure~\ref{fig:fid_steps_comparison}(a), we have conclusion that \textbf{first}, 0-Reflow which is vanila Rectified flow or FM, cannot allow 1 or 2 steps sampling while 1-RF or bigger-RF can; \textbf{Second}, our RA Reflow method effectively avoids model collapse, resulting in a more efficient training process \ref{fig:fid_steps_comparison}(b). Specifically, Table~\ref{tab:m} indicates that Rectified Flow trained with RA Reflow produces high-quality images using only a few sampling steps, highlighting the straighter flow achieved. Detailed experimental settings and additional ablation studies are provided in Appendix~\ref{app:RCA_setup}.
% We conduct a series of experiments on CIFAR-10 to understand the effect of various hyperparameters on the performance of Rectified Flow trained with RA Reflow. We focus on the mixing ratio $\lambda$, which controls the proportion of synthetic data to real data, and the parameter $\alpha$, which dictates the regeneration frequency of real reverse image-noise pairs. Detailed setting up can be found in Appendix~\ref{app:RCA_setup} As visualized in Figure~\ref{fig:fid_steps_comparison}, our RA Reflow effectively avoids model collapse. Table~\ref{tab:m} shows that Rectified Flow trained with RA Reflow achieves high-quality generation using only a few sampling steps, indicating a straighter flow.
\vspace{-1mm}
% \textcolor{blue}{
\textbf{ORA/ORAS Reflow.} RA Reflow can be considered a pseudo-online method. For ORA, we employ a full-size U-Net using the same settings as in \citet{lipman2022flow, dao2023flow}. As shown in Table~\ref{tab:cifar10} and Table~\ref{tab:celeba}, both ORA and ORAS outperform vanilla Reflow, achieving better FID scores than our RCA method, without requiring additional storage.
% \textbf{Online Collapse-Avoiding Reflow.} RA Reflow can be seen as a pseudo-online method. For OCAR, we use a full-size U-Net with the same settings as in \citet{lipman2022flow, dao2023flow}. Table~\ref{tab:m} Shows the ORA and ORAS achieve the best performance than the vanilla Reflow and even achieve better fid values than our RCA without requiring additional storage. 
% }

\begin{table}[t]
\centering
\tiny
% \scriptsize
\begin{tabular}{lcccc}
\toprule
\multirow{2}{*}{} & \multicolumn{4}{c}{CIFAR10 (32 $\times$ 32)} \\
& 10 NFE & 20 NFE & 50 NFE & Best NFE \\
\midrule
0-RF (ICFM) & \textbf{14.16} & 9.88 & 6.30 & 4.02/152  \\
FM   & 16.00 & 10.70 & 7.76 & 6.12/158 \\
OTCFM & 14.47 & \textbf{9.38} & \textbf{5.78} & \textbf{3.96/134} \\
\midrule
1-RF & 10.83 & 9.75 & 7.49 & 5.95/108  \\
1-RF-RA \textbf{(Ours)} & \textbf{8.68} & \textbf{7.47} & \textbf{6.98} & \textbf{5.61/112} \\
\midrule
2-RF & 14.97 & 12.01 & 10.13 & 9.68/107 \\
2-RF-RA \textbf{(Ours)} & \textbf{11.47} & \textbf{9.12} & \textbf{8.58} & \textbf{7.64/102} \\
\midrule
ORA \textbf{(Ours)} & \textbf{7.02} & 6.30 & 5.96 & 4.27/96 \\
ORAS \textbf{(Ours)} & 7.45 & \textbf{6.01} & \textbf{5.19} & \textbf{4.15/94} \\
\bottomrule
\end{tabular}
\caption{\textbf{Comparison on FID score ($\downarrow$) for unconditional generation on CIFAR10.} Full-scale U-Net.}
\vspace{-4mm}
\label{tab:cifar10}
\end{table}

\begin{table}[t]
\centering
\tiny
% \scriptsize
\begin{tabular}{lcccc}
\toprule
\multirow{2}{*}{} & \multicolumn{4}{c}{CelebA-HQ (256 $\times$ 256)} \\
& 10 NFE & 20 NFE & 50 NFE & Best NFE \\
\midrule
FM   & 16.51 & 8.40 & 5.87 & 5.45/89 \\
\midrule
1-RF & 12.04 & 7.34 & 5.76 & 5.73/71 \\
1-RF-RA \textbf{(Ours)} & \textbf{11.39} & \textbf{7.27} & \textbf{5.61} & \textbf{5.57/69} \\
\midrule
2-RF & 13.27 & 8.71 & 7.05 & 6.28/67 \\
2-RF-RA \textbf{(Ours)} & \textbf{12.89} & \textbf{8.50} & \textbf{6.91} & \textbf{6.10/67} \\
\midrule
ORA \textbf{(Ours)} & 10.89 & 7.12 & 5.60 & 5.52/69 \\
ORAS \textbf{(Ours)} & \textbf{10.86} & \textbf{6.99} & \textbf{5.53} & \textbf{5.49/70} \\
\bottomrule
\end{tabular}
\caption{\textbf{Comparison on FID score ($\downarrow$) for unconditional generation on CelebA-HQ.} $\lambda = 0.5$, $\alpha = 2$, DiT-L/2.}
\vspace{-7mm}
\label{tab:celeba}
\end{table}

\vspace{-3mm}
\section{Conclusion}
Our analysis reveals that Reflow, despite reducing sampling steps, can face MC when repeatedly trained on self-generated data. By examining DAEs, we identified the root causes of this degradation and proved that adding real data mitigate collapse. Building on these insights, we introduced RA Reflow and its variants, which integrate real images via reverse processes to mitigate MC. Experiments on both Gaussian and real datasets confirm that our methods mitigate MC and produce high-quality samples with fewer sampling steps. 

\newpage
\section*{Impact Statement}
Our work on improving generative models can benefit applications such as data augmentation, image restoration, and artistic creation. However, it may also exacerbate risks associated with deceptive content, including deepfake generation. For a detailed survey of deepfake creation and detection, we refer readers to \citet{mirsky2021creation}. Since our approach can produce high-quality synthetic data with fewer sampling steps, its societal impact mirrors that of general generative technologies, underscoring the need for continued research on detection methods and responsible deployment.

% \bibliography{iclr2025_conference}

% \bibliographystyle{iclr2025_conference}
\bibliographystyle{icml2025}
\bibliography{icml2025_conference}

%%%%%%%%%%%%%%%%%%%%%%%%%%%%%%%%%%%%%%%%%%%%%%%%%%%%%%%%%%%%%%%%%%%%%%%%%%%%%%%
%%%%%%%%%%%%%%%%%%%%%%%%%%%%%%%%%%%%%%%%%%%%%%%%%%%%%%%%%%%%%%%%%%%%%%%%%%%%%%%
% APPENDIX
%%%%%%%%%%%%%%%%%%%%%%%%%%%%%%%%%%%%%%%%%%%%%%%%%%%%%%%%%%%%%%%%%%%%%%%%%%%%%%%
%%%%%%%%%%%%%%%%%%%%%%%%%%%%%%%%%%%%%%%%%%%%%%%%%%%%%%%%%%%%%%%%%%%%%%%%%%%%%%%
\newpage
% \appendix
\onecolumn
{\huge \bfseries Appendix}

\etocdepthtag.toc{mtappendix}
\etocsettagdepth{mtchapter}{none}
\etocsettagdepth{mtappendix}{subsection}
\tableofcontents
\appendix

% \setcounter{lemma}{0}
% \setcounter{proposition}{0}
% \setcounter{remark}{0}

% 修改编号格式
\renewcommand{\thelemma}{\Alph{section}.\arabic{lemma}}
\renewcommand{\theproposition}{\Alph{section}.\arabic{proposition}}
\renewcommand{\theremark}{\Alph{section}.\arabic{remark}}

\section{Proofs and Formulations} \label{Proofs and Formulations}

\subsection{Proof of Theorem~\ref{prop: DAE collapse}}
\begin{proof}[Proof of \Cref{prop: DAE collapse}] 
% \zz{A very rough sketch. We can move the proof to Appendix.} \mh{Yes, I have copied all the old ones in Appendix~\ref{Proofs and Formulations}}
We can first expand the training loss in \eqref{eq:empirical-DAE-Loss} as follows:
\begin{align}
\mathcal{L}(\theta) &= \mathbb{E}_{\vz \sim \mathcal{N}(0,\sigma^2 I)}\left[ \left\| \mW_2\mW_1 \vx_i - \vx_i \right\|^2 - 2 \langle \mW_2 \mW_1 \vx_i -  \vx_i,  \mW_2 \mW_1 \vz\rangle + \|\mW_2\mW_1 \vz\|_2^2   \right]
,\nonumber\\
& = \sum_{i=1}^n \left\| \mW_2\mW_1 \vx_i - \vx_i \right\|^2 + \sigma^{2}\|\mW_2\mW_1\|_F^2.
\label{regulized version of linear DAE loss}\end{align}
We denote by $\mPhi = \mW_2 \mW_1$ to simplify the following analysis. The induced $\ell_2$ regularization in \eqref{regulized version of linear DAE loss} suggests that DAE performs denoising by learning a low-dimensional model. The optimal solution for \eqref{regulized version of linear DAE loss}, written in terms of $\mPhi$, is simply given by $(\mX\mX^\top)(\mX\mX^\top + \sigma^2 I)^{-1}$. When $\sigma \rightarrow 0$, the solution converges to PCA. Plugging this into the process of recursively learning DAE from generational data, we have $\mathbf{\Phi}_j = \mW_2^j \mW_1^j =  (\mX_j \mX_j^\top)(\mX_j\mX_j^\top +\sigma^2 I)^{-1}$.%, 

Let $\lambda(\cdot)$ denote the largest eigenvalue of a matrix. Since $\mX_{j+1} = \mPhi_j (\mX_j + \mE_j)$ with each column of $\mE_j$ being iid sampled from  $\calN(0,\hat \sigma^2/n^2 I)$, it follows from \cite[Theorem 4.6.1]{vershynin2018high} that there exists a constant $C$ such that, with probability at least $1 - 2e^{-n}$, $ \lambda(\mX_{j+1} \mX_{j+1}^\top) \le \lambda^2(\mPhi_j)(\lambda(\mX_j\mX_j^\top) + C \hat \sigma^2)$. This together with $
\lambda (\mathbf{\Phi}_j) = \frac{\lambda(\mX_j \mX_j^\top)}{\lambda(\mX_j \mX_j^\top) + \sigma^2}$ implies that when $\hat \sigma^2 \le \sigma^2/C$,
\begin{align}
\lambda(\mX_{j+1} \mX_{j+1}^\top) \le \lambda(\mX_j\mX_j^\top) \lambda(\mPhi_j) \frac{\lambda(\mX_j\mX_j^\top) + C \hat \sigma^2}{\lambda(\mX_j\mX_j^\top) + \sigma^2} \le \lambda(\mX_j\mX_j^\top) \lambda(\mPhi_j) 
\label{eq:recursive-E}\end{align}
holds with probability at least $1 - 2e^{-n}$. Denote by $\tau = \lambda(\mX\mX^\top)$. 
In the following, we prove that with probability at least $1 - 2q e^{-n}$, 
\begin{align}
\left[ \lambda(\mX_{q} \mX_{q}^\top) \right] \le \lambda(\mX\mX^\top)(\frac{\tau}{\tau + \sigma^2})^{q-1}.
\label{eq:eig-decay}\end{align}
We prove this by induction. It holds when $q = 0$. Now assume \eqref{eq:eig-decay} is true at $q = j$. We prove it also holds at $q = j+1$. 
Since \eqref{eq:eig-decay} holds at $j$, we have $\lambda(\mX_j \mX_j^\top) \le \lambda(\mX \mX^\top)$, and hence
$\lambda(\mPhi_j) = \frac{\lambda(\mX_j \mX_j^\top)}{\lambda(\mX_j \mX_j^\top) + \sigma^2} \le \frac{\tau}{\tau + \sigma^2}$.
Plugging this into \eqref{eq:recursive-E} gives
\[
\lambda(\mX_{j+1} \mX_{j+1}^\top)  \le \lambda(\mX_j\mX_j^\top) \lambda(\mPhi_j) \le \lambda(\mX\mX^\top)(\frac{\tau}{\tau + \sigma^2})^{j}.
\]
This proves \eqref{eq:eig-decay}. Finally, we can obtain the bound for $\lambda(\mPhi_j)$ as
\[
\lambda(\mPhi_j) = \frac{\lambda(\mX_j \mX_j^\top)}{\lambda(\mX_j \mX_j^\top) + \sigma^2} \le \frac{\lambda(\mX\mX^\top)(\frac{\tau}{\tau + \sigma^2})^{j-1}}{\lambda(\mX\mX^\top)(\frac{\tau}{\tau + \sigma^2})^{j-1} + \sigma^2} \le \frac{\lambda(\mX\mX^\top)}{\sigma^2} (\frac{\tau}{\tau + \sigma^2})^{j-1}.
\]

\end{proof}

\subsection{Detailed Explanation of the Gap Between Diffusion Models and DAEs} \label{remark: DAE and diffusion model}
In this appendix, we delve deeper into the connection between diffusion models and sequences of Denoising Autoencoders (DAEs), focusing on the initial step of the diffusion process.

Consider a diffusion model $f_{\theta}(t, \vx_t)$ with $T$ time steps (e.g., $T=1000$), which begins the sampling process from pure Gaussian noise $\vx_0 \sim \mathcal{N}(0, \mathbf{I})$. The model predicts the target state using (here we consider the image $x$-prediction which is equal to noise $\epsilon$-prediction and velocity $v$-prediction \citep{salimans2022progressive}): 
\begin{equation} 
    \vx_1 = f_{\theta}(0, \vx_0), 
\end{equation} 
where $f_{\theta}(0, \vx_0)$ approximates the denoising function at time $t=0$. This step functions as a DAE with pure Gaussian input.

Subsequent sampling steps involve Euler updates of the form: \begin{equation} 
\begin{aligned}
    & \vx_{0+\gamma} = \vx_0 + \gamma \left( f_{\theta}(0, \vx_0) - \vx_0 \right) \\
    & \cdots \\
    & \vx_{t+\gamma} = \vx_t + \gamma \left( f_{\theta}(t, \vx_t) - \vx_t \right),  
\end{aligned}
\end{equation} 
where $\gamma$ is a small time increment. In these steps, each input $\vx_t$ is a mixture of Gaussian noise and previous model outputs, aligning with the typical input to a DAE trained on such mixtures.

The only significant gap between a sequence of DAEs and the diffusion model arises in the initial step due to the pure Gaussian input. By analyzing the initial step separately, we can better align the recursive DAE framework with the diffusion model. Specifically, if we consider the initial DAE handling pure Gaussian inputs and subsequent DAEs processing mixtures of noise and signal, the entire diffusion process can be viewed as a series of DAEs with varying input distributions.

However, an important question arises: \textit{Will a linear DAE learn any meaningful information from the first step with pure Gaussian input?} In the case of a linear DAE, learning from pure noise is challenging because there is no underlying structure to capture. This limitation highlights why the initial step differs from the rest and underscores the necessity of separating its analysis.

By acknowledging this gap, our analysis becomes more comprehensive, bridging the understanding between DAEs and diffusion models. This perspective not only sheds light on the mechanics of diffusion models but also provides a pathway for leveraging insights from DAE analysis to improve diffusion-based generative models.

\subsection{Proof of Proposition 2}

Now, we formulate the reflow process of a linear DAE incorporating real data. Recall the settings from \ref{Setting: Synthetic Data Generation Process}; suppose we have training data \(\mX = \begin{bmatrix} \vx_1 & \cdots & \vx_n \end{bmatrix}\) with \(\vx_i = \mU^\star \mU^{\star\top} \va_i\), where \(\va_i \sim \mathcal{N}(0, \mathbf{I})\). Starting with \(\mX_1 = \mX\), the scheme for generating synthetic data at the \(j\)-th iteration (\(j \geq 1\)) is outlined as follows.

\phantomsection 
\label{meth: Reflow of DAE with real data}
\begin{itemize}
\item \textbf{Add real data}: \(\hat{\mX}_j = \begin{bmatrix} \mX_j & \mX \end{bmatrix}\).
\item \textbf{Fit DAE}: \((\mW_2^j, \mW_1^j) = \theta^\star(\hat{\mX}_j)\) by solving \eqref{eq:empirical-DAE-Loss} with training data \(\hat{\mX}_j\).
\item \textbf{Generate synthetic data for the next iteration}: \(\mX_{j+1} = \mW_2^{j} \mW_1^j (\mX_j + \mE_j)\), where each column of the noise matrix \(\mE_j\) is i.i.d. sampled from \(\mathcal{N}(0, \hat{\sigma}^2 / n^2 \mathbf{I})\).
\end{itemize}

First, we examine the effect of incorporating real data into the training process. Let $\lambda(\cdot)$ denote the largest eigenvalue of a matrix and $\lambda_{\min}(\cdot)$ denote the smallest eigenvalue of a matrix.

\begin{tcolorbox}[colback=gray!10, colframe=white, sharp corners, boxrule=0pt]
\begin{lemma}
\label{Lemma:add real data}
Let \( X_j, X_0 \in \mathbb{R}^{n \times d} \) be given matrices, and define the block matrix.
\[
\hat{\mX_j} = \begin{bmatrix} X_j &  X_0 \end{bmatrix}.
\]
Then the maximum eigenvalue of \( \hat{\mX_j}\hat{\mX_j}^\top \)
satisfies the following inequalities:
\[
\lambda_{\min}(X_j X_j^\top) + \lambda(X_0 X_0^\top)\leq \lambda(\hat{\mX_j}\hat{\mX_j}^\top) \leq \lambda(X_j X_j^\top) + \lambda(X_0 X_0^\top).
\]\label{eq:lemma_inequality}
\end{lemma}
\end{tcolorbox}

\begin{proof}
First, observe that
\begin{equation}
\label{eq:sum_matrices}
\hat{X}_j \hat{X}_j^\top = X_j X_j^\top + X_0 X_0^\top.
\end{equation}

We aim to bound \( \lambda(\hat{X}_j \hat{X}_j^\top) \) using the eigenvalues of \( X_j X_j^\top \) and \( X_0 X_0^\top \). Recall that both \( X_j X_j^\top \) and \( X_0 X_0^\top \) are symmetric positive semi-definite matrices.

\textbf{Upper Bound:}

Using Weyl's inequality for eigenvalues of Hermitian matrices, we have
\begin{equation}
\label{eq:weyl_upper}
\lambda(A + B) \leq \lambda(A) + \lambda(B),
\end{equation}
where \( A \) and \( B \) are symmetric matrices.

Applying this to \( A = X_j X_j^\top \) and \( B = X_0 X_0^\top \), we obtain
\[
\lambda(\hat{X}_j \hat{X}_j^\top) \leq \lambda(X_j X_j^\top) + \lambda(X_0 X_0^\top).
\]

\textbf{Lower Bound:}

Similarly, Weyl's inequality provides a lower bound:
\begin{equation}
\label{eq:weyl_lower}
\lambda(A + B) \geq \lambda_{\min}(A) + \lambda(B).
\end{equation}

Applying this to \( A = X_j X_j^\top \) and \( B = X_0 X_0^\top \), we have
\[
\lambda(\hat{X}_j \hat{X}_j^\top) \geq \lambda_{\min}(X_j X_j^\top) + \lambda(X_0 X_0^\top).
\]

Combining the upper and lower bounds from Equations~\eqref{eq:weyl_upper} and \eqref{eq:weyl_lower}, we establish the inequalities in Equation~\eqref{eq:lemma_inequality}, thus proving the lemma.
\end{proof}

% \begin{tcolorbox}[colback=gray!10, colframe=white, sharp corners, boxrule=0pt]
% \begin{proposition} \label{prop: DAE no collapse}
% In the above synthetic data generation process~\ref{meth: Reflow of DAE}, suppose that the variance of the added noise \( \sigma^2 > 0 \). Then the learned DAE does not suffer from model collapse, and the largest eigenvalue of $\mathbf{\Phi}_j = \mW_2^j \mW_1^j$ satisfies
% \begin{equation}
% \lambda(\mathbf{\Phi}_j) \geq \frac{\lambda(\mX \mX^\top)}{\lambda(\mX_j \mX_j^\top) + \lambda(\mX \mX^\top) + \sigma^2}.
% \end{equation}
% Furthermore, if $\lambda(\mX_j \mX_j^\top)$ is bounded, then $\lambda(\mathbf{\Phi}_j)$ is bounded below by a positive constant.
% \end{proposition}
% \end{tcolorbox}

\begin{tcolorbox}[colback=gray!10, colframe=white, sharp corners, boxrule=0pt]
\begin{proposition} \label{prop: DAE no collapse}
In the above synthetic data generation process~\ref{meth: Reflow of DAE} with adding real data, suppose that the variance of the added noise is not too large, i.e., $\hat \sigma \le C \sigma$ for some universal constant $C$. Then, with probability at least $1-2je^{-n}$,  the learned DAE does not suffer from model collapse as
\begin{equation}
\|\mW_2^j \mW_1^j\|^2 \geq \frac{\|\mX\|^2}{2\|\mX\|^2 + \sigma^2}.
\end{equation}
\end{proposition}
\end{tcolorbox}

\begin{proof}
Following an analysis similar to the proof of Theorem~\ref{prop: DAE collapse}, we have
\begin{equation}
\label{eq:phi_expression}
\mathbf{\Phi}_j = (\hat{\mathbf{X}}_j \hat{\mathbf{X}}_j^\top) \left( \hat{\mathbf{X}}_j \hat{\mathbf{X}}_j^\top + \sigma^2 \mathbf{I} \right)^{-1},
\end{equation}
where $\hat{\mathbf{X}}_j = \begin{bmatrix} \mathbf{X}_j & \mathbf{X} \end{bmatrix} \in \mathbb{R}^{n \times 2d}$. Since both $\mathbf{\Phi}_j$ and $\hat{\mathbf{X}}_j \hat{\mathbf{X}}_j^\top$ are symmetric positive semi-definite matrices, their eigenvalues are real and non-negative. Therefore, the eigenvalues of $\mathbf{\Phi}_j$ satisfy
\begin{equation}
\lambda(\mathbf{\Phi}_j) = \frac{\lambda(\hat{\mathbf{X}}_j \hat{\mathbf{X}}_j^\top)}{\lambda(\hat{\mathbf{X}}_j \hat{\mathbf{X}}_j^\top) + \sigma^2}.
\end{equation}

Applying the eigenvalue bounds from Lemma~\ref{Lemma:add real data}, we obtain
\begin{align}
\lambda_{\min}(\hat{\mathbf{X}}_j \hat{\mathbf{X}}_j^\top) &\geq \lambda_{\min}(\mathbf{X}_j \mathbf{X}_j^\top) + \lambda_{\min}(\mathbf{X} \mathbf{X}^\top), \label{eq:hatX_lower_bound} \\
\lambda(\hat{\mathbf{X}}_j \hat{\mathbf{X}}_j^\top) &\leq \lambda(\mathbf{X}_j \mathbf{X}_j^\top) + \lambda(\mathbf{X} \mathbf{X}^\top). \label{eq:hatX_upper_bound}
\end{align}

Substituting these bounds into the expression for $\lambda_{\min}(\mathbf{\Phi}_j)$, we have
\begin{equation}
\lambda(\mathbf{\Phi}_j) \geq \frac{\lambda_{\min}(\mathbf{X}_j \mathbf{X}_j^\top) + \lambda(\mathbf{X} \mathbf{X}^\top)}{\lambda(\mathbf{X}_j \mathbf{X}_j^\top) + \lambda(\mathbf{X} \mathbf{X}^\top) + \sigma^2}.
\end{equation}

Since $\lambda_{\min}(\mathbf{X}_j \mathbf{X}_j^\top) \geq 0$, it follows that
\begin{equation}\label{eq:phi_eigenvalue_inequality}
\lambda(\mathbf{\Phi}_j) \geq \frac{\lambda(\mathbf{X} \mathbf{X}^\top)}{\lambda(\mathbf{X}_j \mathbf{X}_j^\top) + \lambda(\mathbf{X} \mathbf{X}^\top) + \sigma^2}.
\end{equation}

Let us denote $\tau = \lambda(\mathbf{X} \mathbf{X}^\top)$ and assume that $\lambda(\mathbf{X}_j \mathbf{X}_j^\top) \leq \tau$ (we will justify this assumption later). Then, we have
\[
\lambda(\mathbf{\Phi}_j) \geq \frac{\lambda(\mathbf{X} \mathbf{X}^\top)}{2\tau + \sigma^2}.
\]

Using a similar analysis as in the proof of Theorem~\ref{prop: DAE collapse}, and the fact that $\mathbf{X}_{j+1} = \mathbf{\Phi}_j (\mathbf{X}_j + \mathbf{E}_j)$, where each column of $\mathbf{E}_j$ is independently sampled from $\mathcal{N}\left(0, \frac{\hat{\sigma}^2}{n^2} \mathbf{I}\right)$, we have
\begin{align}
\lambda(\mathbf{X}_{j+1} \mathbf{X}_{j+1}^\top) \leq \lambda^2(\mathbf{\Phi}_j) \left( \lambda(\mathbf{X}_j \mathbf{X}_j^\top) + C \hat{\sigma}^2 \right),
\label{eq:inequality-Xj}
\end{align}
with probability at least $1 - 2e^{-n}$.

We will now prove that, with probability at least $1 - 2q e^{-n}$, the following holds:
\begin{align}
\lambda(\mathbf{X}_{q} \mathbf{X}_{q}^\top) \leq \tau.
\label{eq:eig-decay2}
\end{align}
We proceed by induction. For $q = 0$, the inequality holds by the definition of $\tau$. Assume that inequality~\eqref{eq:eig-decay2} holds for $q = j$; we will show it also holds for $q = j+1$.

Since \eqref{eq:eig-decay2} holds at iteration $j$, we have $\lambda(\mathbf{X}_j \mathbf{X}_j^\top) \leq \tau$. Therefore,
\[
\lambda(\mathbf{\Phi}_j) \leq \frac{\lambda(\mathbf{X}_j \mathbf{X}_j^\top) + \lambda(\mathbf{X} \mathbf{X}^\top)}{\lambda(\mathbf{X}_j \mathbf{X}_j^\top) + \lambda(\mathbf{X} \mathbf{X}^\top) + \sigma^2} \leq \frac{2\tau}{2\tau + \sigma^2}.
\]

Plugging this bound, along with the assumption $\hat{\sigma}^2 \leq \frac{\sigma^2}{2C}$, into inequality~\eqref{eq:inequality-Xj}, we obtain
\[
\lambda(\mathbf{X}_{j+1} \mathbf{X}_{j+1}^\top) \leq \left( \frac{2\tau}{2\tau + \sigma^2} \right)^2 \left( \tau + C \hat{\sigma}^2 \right) \leq \tau.
\]
This completes the induction step and proves inequality~\eqref{eq:eig-decay2}.

Recall the inequality~\eqref{eq:phi_eigenvalue_inequality}:
$$
\lambda(\mathbf{\Phi}_j) \geq \frac{\lambda(\mathbf{X} \mathbf{X}^\top)}{\lambda(\mathbf{X}_j \mathbf{X}_j^\top) + \lambda(\mathbf{X} \mathbf{X}^\top) + \sigma^2}.
$$
Since $\lambda(\mathbf{X}_j \mathbf{X}_j^\top)$ is bounded above by $\tau$ and $\lambda(\mathbf{X} \mathbf{X}^\top) > 0$, the right-hand side of inequality~\eqref{eq:phi_eigenvalue_inequality} is bounded below by a positive constant. Therefore, $\lambda(\mathbf{\Phi}_j)$ is bounded below by a positive constant, which implies that the learned DAE does not suffer from model collapse.
\end{proof}

\begin{remark}
    To prevent model collapse in generative models, a common strategy is to incorporate real data into the training process. Previous studies \citep{bertrand2023stability, alemohammad2023selfconsuming, gerstgrasser2024model} have shown that mixing real data with synthetic data during training helps maintain model performance and prevents degeneration caused by relying solely on self-generated data. In diffusion models, integrating real samples can enhance model performance and reduce the risk of collapse \citep{kim2023consistency}. By conditioning the model on both real and synthetic data, the training process leverages the structure of real data distributions. Building on these approaches, our work introduces methods to integrate real data into the training of Rectified Flow, even when direct noise-image pairs are not available. By generating noise-image pairs from real data using reverse processes and balancing them with synthetic pairs, we prevent model collapse while retaining efficient sampling.
\end{remark}

\subsection{Model collapse in Rectified flow}\label{app:diss_for_MC_RF}
In the appendix, we provide the formal statement of our proposition and the detailed proof:

\begin{proof}\label{prof: Model Collapse in Rectified Flow}
Consider the explicit Euler discretization of the Rectified Flow ODE. Starting from $\vx_{j,0} = \vz$, where $\vz \sim \mathcal{N}(0, \mathbf{I})$, we update:

\begin{equation} \vx_{j,t+\gamma} = \vx_{j,t} + \gamma, v_{\theta_j}(t, \vx_{j,t}), \quad t \in [0, 1], \end{equation}

with step size $\gamma$. If each small step of $v_{\theta_j}$ acts similarly to a DAE, then based on Theorem~\ref{prop: DAE collapse}, as $j \to \infty$, we have:

\begin{equation}\label{eq
} \lim_{j \to \infty} \operatorname{rank}(v_{\theta_j}) = 0. \end{equation}

This implies $v_{\theta_j}(t, \vx_{j,t}) \to \mathbf{0}$, leading to $\vx_{j,t+\gamma} \approx \vx_{j,t}$. Thus, the generated result remains near the initial point, confirming model collapse as stated in Proposition~\ref{prop:MCRF}.
\end{proof}
\begin{remark} Although there is a theoretical gap between DAEs and Rectified Flow, our experimental results (Figure~\ref{fig:RF_linear_curve}) support this proposition, suggesting that model collapse does occur in Rectified Flow under iterative self-training. \end{remark}

\section{Methods Details}
\subsection{Real-data Augmented Reflow}
\begin{algorithm}[t]
    \caption{Real-data Augmented Reflow}\label{algo:RCA}
    \begin{algorithmic}[1]
        \Require Reflow iterations $\mathcal{J}$; real dataset $\{\vx^{(i)}\}$; pre-trained vector field $v_{\theta_0}$; mix ratio $\lambda$; ODE solver $\mathrm{ODE}_{v_{\theta_0}}(t_0, t_1, \vx)$; regeneration parameter $\alpha$.
        \Ensure Trained vector fields $\{v_{\theta_j}\}_{j=1}^{\mathcal{J}}$
        \For{$j = 1$ to $\mathcal{J}$}
            \State Sample $\{\vz^{(i)}\}$ from $\mathcal{N}(\mathbf{0}, \mathbf{I})$ 
            \State Compute $\hat{\vx}^{(i)} = \mathrm{ODE}_{v_{\theta_j}}(0, 1, \vz^{(i)})$
            % \zz{miss $v_{\theta_j}$?} 
            \Comment{Generate synthetic noise-image pairs}
            \State Compute $\hat{\vz}^{(i)} = \mathrm{ODE}_{v_{\theta_j}}(1, 0, \vx^{(i)})$ 
            \Comment{Generate reverse image-noise pairs from real data}
            
            \State Randomly select $\lambda n$ synthetic pairs and $(1 - \lambda) n$ real reverse pairs
            \State $\mathcal{D}_j = \{ (\vz_j^{(i)}, \vx_j^{(i)}) \}_{i=1}^n = \{ (\vz^{(i)}, \hat{\vx}^{(i)}) \}_{i=1}^{\lambda n} \cup \{ (\hat{\vz}^{(i)}, \vx^{(i)}) \}_{i=1}^{(1 - \lambda) n}$ 
            \Comment{Mix Pairs with Ratio $\lambda$}
            
            \Repeat 
            \Comment{Reflow training}
                \For{each $(\vz_j^{(i)}, \vx_j^{(i)}) \in \mathcal{D}_j$}
                    \State Sample $t \sim \mathcal{U}(0, 1)$
                    \State Compute $\vx_t^{(i)} = t\, \vx_j^{(i)} + (1 - t)\, \vz_j^{(i)}$
                    \State Compute loss:
                    \[
                        \mathcal{L}_{\mathrm{RF}} = \frac{1}{B} \sum_{i=1}^{B} \left\| v_{\theta_j}(t, \vx_t^{(i)}) - (\vx_j^{(i)} - \vz_j^{(i)}) \right\|^2
                    \]
                    \State Update $\theta_j$ using gradient descent
                \EndFor
                \State{Repeat Steps 4 and 6 every $\alpha$ epoches}
                % \If{$ j \mod \alpha = 0 $}
                %     % \Comment{Re-generate pairs every $\alpha$ epochs}
                % \State Repeat Steps 4 and 6 
                % \EndIf
            \Until{converged}
                    % \zz{the outside for loop, inside repeat loop, and inside if loop seems not correct; please double check.}
        \EndFor
        \State \textbf{Output}: $\{v_{\theta_j}\}_{j=1}^{\mathcal{J}}$
    \end{algorithmic}
\end{algorithm}

\subsection{Online Real-data Augmented Reflow} \label{algo: OCAR-datail}
\begin{algorithm}[h]
    \caption{Online Real-data Augmented Reflow Training}\label{algo:OCAR}
    \begin{algorithmic}[1]
        \Require Reflow iterations $\mathcal{J}$; real dataset $\{\vx^{(i)}\}$; pre-trained vector field $v_{\theta_0}$; mix ratio $\lambda$; SDE/ODE solver $\mathrm{SDE/ODE}(t_0, t_1, \cdot)$; regeneration parameter $\alpha$
        \Ensure Trained vector fields $\{v_{\theta_j}\}_{j=1}^{\mathcal{J}}$
        \For{$j = 1$ to $\mathcal{J}$}
            \Repeat \Comment{Reflow training}
                \For{each mini-batch}
                    \State Sample $\{\vz^{(i)}\}$ from $\mathcal{N}(\mathbf{0}, \mathbf{I})$
                    \State Compute $\hat{\vx}^{(i)} = \mathrm{SDE/ODE}(0, 1, \vz^{(i)})$ \Comment{Generate synthetic data}
                    \State Sample $\{\vx^{(i)}\}$ from real dataset
                    \State Compute $\hat{\vz}^{(i)} = \mathrm{SDE/ODE}(1, 0, \vx^{(i)})$ \Comment{Generate reverse data}
                    \State Randomly select $\lambda B$ synthetic pairs and $(1 - \lambda) B$ real reverse pairs
                    \State $\mathcal{D}_j = \{ (\vz_j^{(i)}, \vx_j^{(i)}) \} = \{ (\vz^{(i)}, \hat{\vx}^{(i)}) \} \cup \{ (\hat{\vz}^{(i)}, \vx^{(i)}) \}$ \Comment{Mix pairs according to $\lambda$}
                    \State Sample $t \sim \mathcal{U}(0, 1)$
                    \For{each $(\vz_j^{(i)}, \vx_j^{(i)})$ in $\mathcal{D}_j$}
                        \State Compute $\vx_t^{(i)} = t\, \vx_j^{(i)} + (1 - t)\, \vz_j^{(i)}$
                        \State Compute loss:
                        \[
                            \mathcal{L}_{\mathrm{RF}} = \frac{1}{B} \sum_{i=1}^{B} \left\| v_{\theta_j}(t, \vx_t^{(i)}) - (\vx_j^{(i)} - \vz_j^{(i)}) \right\|^2
                        \]
                    \EndFor
                    \State Update $\theta_j$ using gradient descent
                \EndFor
            \Until{converged}
        \EndFor
        \State \textbf{Output}: $\{v_{\theta_j}\}_{j=1}^{\mathcal{J}}$
    \end{algorithmic}
\end{algorithm}

\subsection{Does Adding Randomness Help? Reverse SDE Sampling}
\label{meth:S}

In the previous methods, we utilized the reverse ODE process to generate noise-image pairs for training. However, when using only the deterministic ODE, the randomness in the training process originates solely from the initial latent variables $\vz \sim \mathcal{N}(\mathbf{0}, \mathbf{I})$. This limited source of randomness may impact the diversity of the generated samples and the robustness of the model \citep{zhang2023emergence}.

To enhance diversity and potentially improve generation quality, we consider introducing controlled randomness into the reverse process by employing a reverse Stochastic Differential Equation (SDE). The reverse SDE allows us to inject noise at each time step during the sampling process, defined as:

\begin{equation}\label{eq
} d\vx = \left[ f(t, \vx) - g(t)^2 \nabla_{\vx} \log p_t(\vx) \right] dt + g(t) d\widetilde{\mathbf{w}}, \end{equation}

where $f(t, \vx)$ and $g(t)$ are the drift and diffusion coefficients, respectively, and $d\widetilde{\mathbf{w}}$ denotes the standard Wiener process in reverse time. By introducing the diffusion term $g(t) d\widetilde{\mathbf{w}}$, we inject controlled stochasticity into the reverse sampling.

In practice, we set the noise scale $g(t)$ to be small (e.g., $\sigma = 0.001$) and perform sampling using methods like the Euler-Maruyama scheme with an appropriate number of steps (e.g., 100 steps). This controlled injection of noise increases the variability in the training data without significantly disrupting the straightening effect of the flow. Specifically, the added randomness helps explore the neighborhood of data samples, enriching the learning process. We denote this method as \textbf{ORS Refflow}.
\section{Experiments Details and Extra Results}
\subsection{Gaussian Task}\label{app:Gaussian}
\textbf{Setup for DAE.} In the Reflow verification experiment for the DAE, we use a 4-dimensional Gaussian distribution as both the initial and target distributions. The initial distribution is $\mathcal{N}(\mathbf{0}, \mathbf{I})$, and the target distribution is $\mathcal{N}(\mathbf{0}, 5\mathbf{I})$, where $\mathbf{0}$ is a 4-dimensional zero vector, and $\mathbf{I}$ is the identity matrix. We employ a neural network $\theta$ composed of two linear layers $\mW_1$ and $\mW_2$ without activation functions and biases. We train the Reflow process for 20 iterations. The "Ratio" refers to the proportion of synthetic data to real data; a higher value indicates a greater proportion of synthetic data.

Figure~\ref{fig:DAE_linear_curve} presents the results from the Reflow experiment using a Denoising Autoencoder (DAE) on a 4-dimensional Gaussian distribution.

\textbf{(a)} illustrates the rank of the weight matrix \( \mW \) across different Reflow iterations. We set a threshold of $2 \times 10^{-1}$. Specifically, we perform Singular Value Decomposition (SVD) on \( \mW_j \) and count the number of singular values greater than or equal to $0.2$ to determine the rank of \( \mW \). The results demonstrate that incorporating real data effectively prevents model collapse, as indicated by the maintenance of higher ranks. In contrast, relying solely on self-generated synthetic data leads to a rapid decline in rank towards zero.

\textbf{(b)} shows the Wasserstein-2 (W2) distance between the true target data distribution and the generated data distribution over Reflow iterations. This metric assesses the fidelity of the generated data in approximating the target distribution.

\textbf{(c)} displays the evolution of the first principal component (Dimension 0) of the data as Reflow iterations increase. We compare the original DAE, which does not utilize synthetic data, with our DAE-CA model, which employs various ratios of synthetic data (ranging from 0.1 to 0.9), as well as a fully synthetic DAE. The comparison highlights the effectiveness of our DAE-CA model in maintaining the integrity of principal components, thereby preserving data structure and diversity.

\textbf{Setup for Rectified Flow.} In the Reflow verification experiment for linear neural network Rectified Flow, we augment $\mW_1$ by adding one dimension corresponding to time, resulting in a neural network $\mW_1\mW_2: \mathbb{R}^{d+1} \to \mathbb{R}^d$. Our experimental results can be found in the below and confirm our prop \ref{prop:MCRF}
We also test a nonlinear neural network consisting of three linear layers with SELU activation functions and an extra dimension added to the first linear layer. The results are shown in 

\subsection{Model collapse in linear Rctified Flow}
\begin{figure}[t]
  \centering
  \begin{subfigure}[b]{0.32\linewidth}
    \centering
    \includegraphics[width=\linewidth]{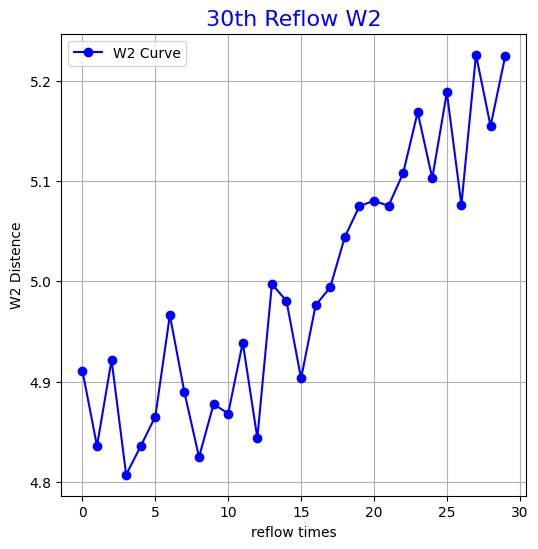}
    \caption{Wasserstain-2 Distance}
    \label{fig:W2RF}
  \end{subfigure}
  \hfill
  \begin{subfigure}[b]{0.32\linewidth}
    \centering
    \includegraphics[width=\linewidth]{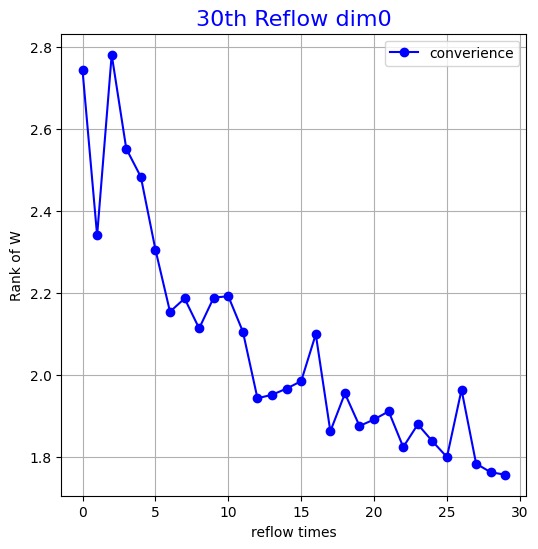}
    \caption{Dim 0}
    \label{fig:dim0RF}
  \end{subfigure}
  \hfill
  \begin{subfigure}[b]{0.32\linewidth}
    \centering
    \includegraphics[width=\linewidth]{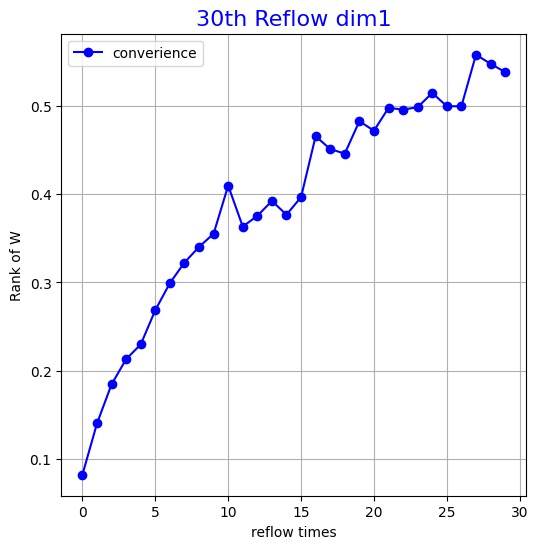}
    \caption{Dim 1}
    \label{fig:dim1RF}
  \end{subfigure}
  \caption{
    Results from the reflow experiment with linear Rectified flow on 10D Gaussian. 
}
\label{fig:RF_linear_curve}
\end{figure}

We experiment on a 10-dimensional Gaussian which starts from the initial distribution $\mathcal{N}(\mathbf{0}, \mathbf{I})$, and the target distribution is $\mathcal{N}(\mathbf{0}, 5\mathbf{I})$. But to demonstrate our inference, we set dimension 1 of the covariance matrix to 1e-3, which reduces the rank of the data as a whole. Figure~\ref{fig:W2RF} shows the model collapse process of linear RF, the Figure~\ref{fig:dim0RF} and Figure~\ref{fig:dim1RF} demonstrates the correctness of Propositio~\ref{prop:MCRF}

\subsection{Straight Flow and Fewer-Step Image Generation}\label{app:RCA_setup}
\footnote{Our results may differ slightly from those reported in the original papers due to variations in neural network settings or random seeds. Nonetheless, our comparisons are fair. Because model collapse experiments require extensive retraining, we used more resource-efficient settings, which can further contribute to these minor discrepancies.}
\begin{figure}[t]
  \centering
  \includegraphics[width=\linewidth]{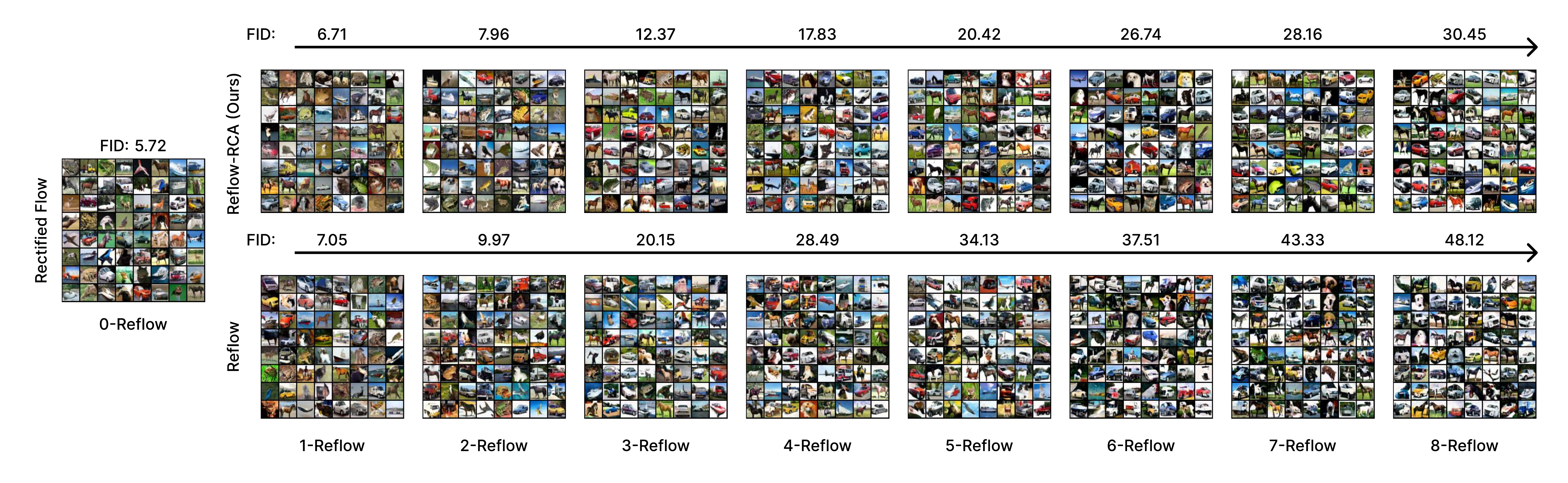}
  \caption{
    Results from the reflow experiment in CIFAR-10 using half-scale U-net.}
    \label{fig: 8-Reflow cifar10}
\end{figure}

In our RCA Reflow experiments, due to the high computational cost of Reflow training, we use a half-size U-Net compared to the one used in Flow Matching \citep{lipman2022flow}. For the qualitative experiments on CIFAR-10 shown in Table~\ref{tab:cifar10}, we use a full-size U-Net with settings consistent with \citet{lipman2022flow} to achieve the best performance. We used the standard implementation from the \url{https://github.com/atong01/conditional-flow-matching} repository provided by \citet{tong2023improving}. All methods were trained using the same configuration, differing only in the choice of the probability path or Reflow methods. Since the code for \citet{lipman2022flow} has not been released, some parameters may still differ from the original implementation. We summarize our setup here; the exact parameter choices can be found in our source code. We used the Adam optimizer with $\beta_1 = 0.9$, $\beta_2 = 0.999$, $\epsilon = 10^{-8}$, and no weight decay. To replicate the architecture in \citet{lipman2022flow}, we employed the U-Net model from \citet{dhariwal2021diffusion} with the following settings: channels set to 256, depth of 2, channel multipliers of [1, 2, 2, 2], number of heads as 4, head channels as 64, attention resolution of 16, and dropout of 0.0. We also used the "ICFM" methods from \citet{tong2023improving}'s repository to train Rectified Flow instead of using the original repository open-sourced by \citet{liu2022flow}, because they use the same interpolation methods and probability paths.

Training was conducted with a batch size of 256 per GPU, using six NVIDIA RTX 4090 GPUs, over a total of 2000 epochs. For Reflow, we generated 500,000 noise-image pairs for every Reflow iteration, according to \citet{liu2022flow}'s blog\footnote{\url{https://zhuanlan.zhihu.com/p/603740431}}. Although \citet{liu2022flow} mention that they use 40,00,000 noise-image to get the best performance, we keep the regular 500,000 noise-image to save time and training source. The learning rate schedule involved increasing the learning rate linearly from 0 to $5 \times 10^{-4}$ over the first 45,000 iterations, then decreasing it linearly back to 0 over the remaining epochs. We set the noise scale $\sigma = 10^{-6}$. For sampling, we used Euler integration with the \texttt{torchdyn} package and the \textsc{dopri5} solver from the \texttt{torchdiffeq} package.

\begin{table}[ht]
   \centering
   \caption{Summary of Configuration Parameters Across Experiments}
   \begin{tabular}{lccc}
   \toprule
   & \textbf{CIFAR10-figure~\ref{fig:method}} & \textbf{CIFAR10-figure~\ref{fig:fid_steps_comparison}} & \textbf{CIFAR10-Table~\ref{tab:cifar10}}\\
   \midrule
   Channels & 256 & 128 & 256 \\
   Channels multiple & 1,2,2,2 & 1,2,2,2 & 1,2,2,2 \\
   Heads & 4 & 4 & 4\\
   Heads Channels & 64 & 64 & 64 \\
   Attention resolution & 16 & 16 & 16 \\
   Dropout & 0.0 & 0.0 & 0.0 \\
   Effective Batch size & 256 & 256 & 256\\
   GPUs & 6 & 6 & 6 \\
   Noise-image pairs & 100k & 500k & 500k \\
   Reflow Sampler & dopri5 & Euler (100 NFE) & dopri5 \\
   $\alpha$ & 2 & 4 & 2 \\
   $\lambda$ & 0.1 & / & 0.5 \\
   Learning Rate & 2e-4 & 5e-4 & 5e-4 \\
   \bottomrule
   \end{tabular}
   \label{tab:configuration_parameters}
   \end{table}

% \begin{table}[ht] \centering \caption{Summary of Configuration Parameters Across Experiments} \begin{tabular}{lccc} \toprule & \textbf{CIFAR10-Figure~\ref{fig
% }} & \textbf{CIFAR10-Figure~\ref{fig
% }} & \textbf{CIFAR10-Table~\ref{tab
% }}\ \midrule Channels & 256 & 128 & 256 \ Channel Multipliers & [1, 2, 2, 2] & [1, 2, 2, 2] & [1, 2, 2, 2] \ Heads & 4 & 4 & 4\ Head Channels & 64 & 64 & 64 \ Attention Resolution & 16 & 16 & 16 \ Dropout & 0.0 & 0.0 & 0.0 \ Effective Batch Size & 256 & 256 & 256\ GPUs & 6 & 6 & 6 \ Noise-Image Pairs & 50k & 500k & 500k \ Reflow Sampler & dopri5 & Euler (100 NFE) & dopri5 \ $\alpha$ & 2 & 4 & 2 \ $\lambda$ & 0.1 & --- & 0.5 \ Learning Rate & $2 \times 10^{-4}$ & $5 \times 10^{-4}$ & $5 \times 10^{-4}$ \ \bottomrule \end{tabular} \label{tab
% } \end{table}

In the CelebA-HQ experiments, we maintain the image resolution at $256 \times 256$. We utilize a pretrained Variational Autoencoder (VAE) from Stable Diffusion \citep{rombach2022high}, where the VAE encoder reduces an RGB image $\mathbf{x} \in \mathbb{R}^{h \times w \times 3}$ to a latent representation $\mathbf{z} = \mathcal{E}(\mathbf{x})$ with dimensions $\frac{h}{8} \times \frac{w}{8} \times 4$. We used the standard implementation from the LFM repository (\url{https://github.com/VinAIResearch/LFM}) provided by \citet{dao2023flow}. We also used the DiT-L/2 \citep{peebles2023scalable} checkpoint released in \citet{dao2023flow}'s repository as the starting point for our Reflow training. Training was conducted with 4 NVIDIA A800 GPUs. 

For RCA Reflow, we tested $\lambda \in {0.1, 0.3, 0.5, 0.7, 0.9, 1.0}$ with $\alpha = 4$. Note that when $\lambda = 0.0$, we are using 100\% real reverse image-noise pairs, which is not equivalent to the original Reflow of Rectified Flow. Therefore, we train the original Reflow as the baseline. For the regeneration parameter $\alpha$, we fixed $\lambda = 0.5$ and compared $\alpha \in {2, 4, 10, \infty}$, where $\infty$ means we never regenerate new data within a single Reflow training. We evaluated the models using both the adaptive sampler "dopri5" (consistent with \citet{lipman2022flow}) and fixed, low numbers of function evaluations (NFEs) ${10, 20, 50}$ to demonstrate the elimination of model collapse and the maintenance of flow straightness by our method. This allows us to assess both generation quality and sampling efficiency simultaneously.

\subsection{Extra Results}
\textbf{Parameter Ablation}
Here we set the same setting in table~\ref{tab:configuration_parameters} column 2.
\begin{table}[h]
    \centering
    \caption{Performance of RF-RCA Models under Different $\lambda$ Values}
    \label{tab:rf_rca_results}
    \begin{tabular}{lcccccc}
        \toprule
        $\lambda$ & 0.1 & 0.3 & 0.5 & 0.7 & 0.9 & 1.0 \\
        \midrule
        1-RF-RCA & 5.87 & 6.21 & 6.37 & 6.81 & 6.93 & 7.05 \\
        2-RF-RCA & 6.37 & 7.10 & 7.96 & 8.53 & 8.98 & 9.97 \\
        3-RF-RCA & 8.02 & 10.29 & 12.37 & 14.74 & 18.01 & 20.15 \\
        \bottomrule
    \end{tabular}
\end{table}

\begin{table}[h]
    \centering
    \caption{Performance of RF-RCA Models under Different $\alpha$ Values}
    \label{tab:rf_rca_results}
    \begin{tabular}{lcccc}
        \toprule
        $\alpha$ & 2 & 4 & 8 & $\infty$ \\
        \midrule
        1-RF-RCA & 6.09 & 6.37 & 6.70 & 7.05 \\
        2-RF-RCA & 6.92 & 7.10 & 8.14 & 9.97 \\
        3-RF-RCA & 9.71 & 10.29 & 13.37 & 20.15 \\
        \bottomrule
    \end{tabular}
\end{table}

\textbf{Precision and Recall}
Here we set the same setting in table~\ref{tab:configuration_parameters} column 2.
\begin{table}[H]
    \centering
    \caption{Precision and Recall Performance on CIFAR10 and CelebA-HQ Datasets}
    \label{tab:precision_recall_results}
    \begin{tabular}{lcc}
        \toprule
        Precision/Recall & CIFAR10 & CelebA-HQ \\
        \midrule
        0-RF     & 0.652 / 0.594 & 0.863 / 0.610 \\
        1-RF     & 0.667 / 0.556 & 0.857 / 0.514 \\
        1-RF-RCA & 0.658 / 0.587 & 0.859 / 0.549 \\
        2-RF     & 0.673 / 0.528 & 0.872 / 0.436 \\
        2-RF-RCA & 0.661 / 0.563 & 0.867 / 0.501 \\
        \bottomrule
    \end{tabular}
\end{table}

\textbf{1/2 step results for CIFAR10}
\begin{table}[H]
    \centering
    \caption{Performance of RF-RCA models under different NFEs. Original data from the cited papers are provided in brackets when available. We set $\lambda = 0.5$, $\alpha = 2$, and use full-scale U-Net for CIFAR-10.}
    \label{tab:rf_rca_results}
    \begin{tabular}{lcccc}
        \toprule
        $NFE$ & 1 & 2 \\
        \midrule
        0-RF     & 351.79 (378)  & 154.65 \\
        1-RF     & 15.27 (12.21) & 11.49 \\
        2-RF     & 19.27 (8.15) & 17.57 \\
        1-RF-RCA & 12.27 & 10.89 \\
        2-RF-RCA & 16.04 & 14.99 \\
        \bottomrule
    \end{tabular}
\end{table}

\end{document}